# Principle Based Semantics for HPSG

Anette Frank Uwe Reyle
Institute for Computational Linguistics
University of Stuttgart
Azenbergstr. 12, D – 70174 Stuttgart
e-mail: uwe@ims.uni-stuttgart.de

## Abstract

The paper presents a constraint based semantic formalism for HPSG. The advantages of the formlism are shown with respect to a grammar for a fragment of German that deals with (i) quantifier scope ambiguities triggered by scrambling and/or movement and (ii) ambiguities that arise from the collective/distributive distinction of plural NPs. The syntax-semantics interface directly implements syntactic conditions on quantifier scoping and distributivity. The construction of semantic representations is guided by general principles governing the interaction between syntax and semantics. Each of these principles acts as a constraint to narrow down the set of possible interpretations of a sentence. Meanings of ambiguous sentences are represented by single partial representations (so-called U(nderspecified) D(iscourse) R(epresentation) S(tructure)s) to which further constraints can be added monotonically to gain more information about the content of a sentence. There is no need to build up a large number of alternative representations of the sentence which are then filtered by subsequent discourse and world knowledge. The advantage of UDRSs is not only that they allow for monotonic incremental interpretation but also that they are equipped with truth conditions and a proof theory that allows for inferences to be drawn directly on structures where quantifier scope is not resolved.

## 1 Introduction

The semantic analysis of standard HPSG deviates from the familiar Montegovian way to construct semantic representations mainly in that it uses unification to eliminate the need for $\beta$-reduction. Variables are bound to argument positions by the close interplay between syntactic and semantic processing; and the semantics of constituents is determined by the Semantics Principle, which governs the way of unifying the semantics of daughter constituents to build up the semantic value of the phrasal constituent: The CONTENT value is projected from the *semantic head*, which is defined as the syntactic HEAD-DTR in head-comp-structures, but as the ADJ-DRT in head-adjunct structures. It is important to note that the semantic contribution of quantified verb arguments is not completely projected as part of the CONTENT value. The meaning of such NPs splits into the features QUANTS, a list representing the information about quantifier scope, and NUCLEUS, containing the nonquantificational core. In the general case only the NUCLEUS is projected from the semantic head according to the Semantics Principle, while the QUANTS value gets instantiated stepwise in interaction with the quantifier storage mechanism (Cooper Store).[1] The mechanism of Cooper storage is built into HPSG by use of two further attributes, QSTORE and RETRIEVED, both represented as sets of quantifiers. All quantifiers start out in QSTORE by lexical definition. The Semantics Principle defines the inheritance of QSTORE to the higher levels of structure, where they may be taken out of store by an appropriately instantiated RETRIEVED value and then put into the QUANTS value of the CONTENT feature. The order in which the semantic value of quantified NPs is retrieved thereby fixes their relative scope. To analyse sentences with scope ambiguities several parses are thus necessary. Besides the definition of appropriate restrictions to and configurations for applications of RETRIEVED the main problem we face with this kind of analysis is, therefore, to modify the semantics of HPSG such that it yields underspecified representations and not sets of fully specified ones.

The need for underspecified representations is by now widely accepted within computational and theoretical linguistics.[2] To make the results of the ongoing research on underspecified representations available for HPSG we may persue two strategies. According to the first strategy we take the HPSG-style analysis – essentially as it is – and only apply slight modifications to produce underspecified output. The second strategy involves a more radical change as it takes an existing theory of underspecified representations and replaces the HPSG semantics by the construction principles of this theory.

Let us start out with a sketch of the first approach. It will show us where its limitations are and allow us to compare different approaches to underspecification. The first thing to do, when un-specifying HPSG semantics, is to relax the retrieval operation. This must be done in two respects. First, we must allow NP-meanings not to be retrieved at all. This results in their relative scope not being determined. Second, we must accommodate syntactic and semantic restrictions on possible scope relations to be stated by the grammar.[3] Restrictions specifying, for example, that the subject NP must always have wide scope over the other arguments of the verb; or, that the scope of genuinely quantified NPs is clause bounded. The modifications we propose are the following. First, we incorporate the QSTORE feature into the CONTENT feature structure. This makes the NP meanings available even if they are not retrieved form QSTORE. Second, we take the value of the QUANTS feature not to be a "stack" (i.e. by appending new retrieved quantifiers as first elements to QUANTS), but allow any NP meaning that is retrieved at a later stage to be inserted *at any place* in that list. This means that the order of NP meanings in QUANTS fixes the relative scope of these meanings only; it does not imply that they have narrow scope with respect to the NP meaning that will be retrieved next. But this is not yet enough to implement clause boundedness. The easiest way to formulate this restriction is to prohibit projection of quantified NP meanings across bounding nodes. Thus the QSTORE and QUANTS values of a bounding node inherit the quantificational information only of *indefinite* NPs and not of *generalized quantifiers*. To be more precise, let us consider the tree $\beta$ consisting only of the bounding nodes in the syntactic analysis of a sentence $\gamma$. Then the semantic content of $\gamma$ can be associated with nodes of $\beta$ in the following way. For each node $i$ of $\beta$ the attributes QUANTS, QSTORE and NUCLEUS have values $quants_i$, $qstore_i$ and $nucleus_i$. The relative scope between scope bearing phrases of $\gamma$, i.e. between the elements of $\bigcup_i(quants_i \cup qstore_i)$ can then be defined as follows.

[1] Cooper Storage mechanism was introduced by [Cooper] to deal with scopally ambiguous sentences without postulating a syntactic ambiguity. The underlying idea is to associate with each syntactic node not a single meaning (i.e. a formula of Montagues intensional logic), but a set of pairs consisting of a storage and a formula. The storage is a set of NP meanings that may be retrieved at certain positions. Whenever an NP meaning is taken out of store then it is applied to the formula to produce a new formula via ($\beta$-reduction).

[2] See, e.g. [Peters/vanDeemter95] for a recent discussion.

[3] This has to be done also for the standard theory.

- If $Q_1$ and $Q_2$ are in $quants_i$ and $Q_1$ precedes $Q_2$, then $Q_1$ has scope over $Q_2$.
- If $Q_1$ is in $quants_i$ and $Q_2$ in $quants_j$, where $i$ dominates $j$, then $Q_1$ has scope over $Q_2$.
- If $Q_1$ is in $qstore_i$ and not in $qstore_j$, where $i$ dominates $j$, then $Q_1$ has scope over any $Q_2$ in $qstore_j \cup quants_j$ that are not in $qstore_i \cup quants_i$.

The last clause says that any NP $Q_1$ occurring in the clause of level $i$ and that is still in QSTORE has scope over all quantified NPs $Q_2$ occurring in embedded clauses (i.e. clauses of level $j$). But $Q_1$ does not necessarily have scope over any indefinite NP introduced at level $j$.

Those familiar with the work of Alshawi and Crouch [Alshawi/Crouch] might have noticed the similarity of their interpretation mechanism and what we have achieved by our modifications to standard HPSG semantics. The elements of QUANTS play exactly the same role as the instantiated metavariables of Alshawi and Crouch. This means that we could adapt their interpretation mechanism to our partially scoped CONTENT structures. But note that we already have achieved more than they have as we are able to express the clause-boundeness restriction for generalized quantifiers.

We will not go into the details and show how the truth conditions of Alshawi and Crouch have to be modified in order to apply to partially scoped CONTENT structures. We will instead go ahead and work out the limitations of what we called the first strategy. To keep things as easy as possible we restrict ourselves to the case of simple sentences (i.e. to trivial tree structures of QSTORE and QUANTS values that consist of one single node only). In this case the QUANTS value (as well as the instantiation of metavariables) imposes a partial order on the relative scope of quantifiers. Assume we had a sentence with three quantifiers, $Q_1$, $Q_2$ and $Q_3$. Then the possible lenghts of QUANTS values varies from 0 to 3. Lengths 0 and 1 leave the relative scope of $Q_1$, $Q_2$ and $Q_3$ completely underspecified. Values of length 2 say that their first element always has wide scope over the second, leaving all possible choices for the third quantifier. And finally we have the fully specified scoping relations given by values of length 3. There are, however, some possibilities to restrict scope relationships that cannot be represented this way: One cannot, for example, represent the ambiguity that remains if we (or, syntax and semantics) require that $Q_1$ and $Q_2$ must have scope over $Q_3$, but leaves unspecified the relative scope between $Q_1$ and $Q_2$; nor are we able to express a restriction that says $Q_1$ must have scope over both, $Q_2$ and $Q_3$, while leaving the relative scope between $Q_2$ and $Q_3$ unspecified. Retrieving a quantifier $Q_i$ (or starting to calculate the truth value of a sentence by first considering this quantifier) is an operation that takes $Q_i$ and adds it to QUANTS. As QUANTS is a list this amounts to a full specification of the relative scope of $Q_i$ with respect to *all* other elements already contained in QUANTS. This shows that the expressive power of the representation language is too restrictive already for simple sentences. We need to represent *partial* orders of quantifier scope. But we cannot do this by talking about a pair consisting of a quantifier $Q_i$ and a list of quantifiers QUANTS. We must be able to talk about *pairs of quantifiers*. This not only increases expressive power of the representation language, it also allows for the formulation of restrictions to quantifier scope in a declarative *and* natural way. The formalism of UDRSs we introduce in the following section is particularly suited to 'talk' about semantic information contributed by different components of a sentence. It therefore provides a particularly good ground to realise principle based construction of semantic representations. But before we start introducing UDRSs let us, first, make a remark on Cooper Storage and its applicability to the construction of underspecified representations, and, second, give a list of remaining shortcomings of the first strategy approach.

As we have seen the original Cooper Storage meachanism essentially consists in mapping a *set* of NP meanings to a set of *sequences* of meanings by taking a meaning out of the set and appending it as first element to the sequence. Our first generalization was to replace strict appending with arbitrary insertion at any place in the sequence. But we have seen that, for example, inserting $Q_2$ at any place in the sequence $\langle Q_1, Q_3 \rangle$ will result in a linear order between $Q_1$, $Q_2$ and $Q_3$. To be able to represent partial orders of scoping relations we must, therefore, completely give up the idea to deal with scope ambiguities by mapping a set of NP meanings to a sequence of NP meanings. What we must do is to impose a partial order to the set itself. Of course this is nothing else than mapping the set of NP meanings to a set of pairs of NP meanings. And so we could have gone one step further and replaced the list QUANTS by a set of pairs. Saying that $Q_2$ should only have narrow scope with respect to $Q_1$ in our example, but enter no scoping relation with respect to $Q_3$, would then amount to extend the set $\{\langle Q_1, Q_3 \rangle\}$ to $\{\langle Q_1, Q_2 \rangle, \langle Q_1, Q_3 \rangle\}$. We didn't do this because it seemed to us too far away from anything one could still call a 'storage mechanism'. The approach we will present in this paper is built on the idea of constructing a semantic representation by directly imposing a partial order to a set of lexically triggered information bits and/or meanings of phrasal components.

The main shortcomings of HPSG semantics that remain also in the modified version sketched above are the following. First, adjuncts (like quantificational adverbs, modals) and also negation bear the potential to introduce scope ambiguities. In order to treat them by the same mechanism that treats the arguments of the verb their meaning would have to be put into store. This, however, requires further modifications of the Semantics Principle, because the treatment of head-adjunct structures differs essentially from the treatment of other configurations (see [Pollard/Sag], Ch.8).[4] Second, there is no underspecified representation of ambiguities that arise from the distributive/collective distinction of plural NPs (neither within the HPSG framework nor in the CLE[5]). Third, the semantic representation of indefinite NPs must be independent of the context in which they are interpreted. We do not want to switch from a universally quantified interpretation to an existentially quantified one, when we come to disambiguate the ambiguous sentence **Every student who admires a philosopher reads his original writings.** such that **a philosopher** is interpreted specifically. This requirement calls for DRT as underlying semantic formalism.

[4] For a general criticism of the analysis of adjuncts in standard HPSG see [Abb/Maienborn]. Their analysis of adjuncts in HPSG fits neatly into the account of semantics projection to be presented below.

[5] In CLE the resolution of QLFs also involves disambiguation with respect to this kind of ambiguities.

In the sequel of this paper we show how the extension of DRT to UDRT given in [Reyle 93] can be combined with an HPSG-style grammar. The basic idea of the combination being that syntax as well as semantics provide structures of equal right; that the principles internal to the syntactic and semantic level are motivated *only* by the syntactic and semantic theory, respectively; and that mutually constraining relations between syntax and semantics are governed by a separate set of principles that relate syntactic and semantic information appropriately. We will replace the Semantics Principle of standard HPSG versions by a principle which directly reflects the *monotonicity* underlying the interpretation process designed in [Reyle 93]: At any stage of the derivation more details are added to the description of the semantic relations between the various components of the sentence, i.e. the partial representation of any mother node is the union of the partial representations of its daughter nodes plus further constraints derived from the syntactic, semantic and also pragmatic context. Construction and disambiguation of semantic representations is thus a monotonic process. Monotonicity guarantees that the transition from an underspecified representation $r_1$ to a less underspecified (or even fully specified) representation $r_2$ is achieved only by *adding* information. There is thus no need to restructure (parts of) a semantic representation if more information about scope restriction has become available.

In the present paper we will focus only on principles restricting scope ambiguities and ambiguities resulting from plural NPs. The underlying scope theory was developed originally by Frey in [Frey] for arguments of the verb and has then been extended to include adjuncts in [Frey/Tappe]. We give a brief overview of their theory in Section 3. Section 4 introduces the Semantics Principle governing the construction of UDRSs. Sections 5 and 6 extend the fragment of German to scrambling, scope and plurals.

## 2 A Short Introduction to UDRS's

The base for unscoped representations proposed in [Reyle 93] is the separation of information about the structure of a particular semantic form and of the content of the information bits the semantic form combines. In case the semantic form is given by a DRS its structure is given by the hierarchy of subDRSs, that is determined by $\Rightarrow$, $\neg$, $\vee$ and $\Diamond$. We will represent this hierarchy explicitly by the subordination relation $\leq$. The semantic content of a DRS consists of the set of its discourse referents and its conditions.

Let us consider the DRSs (2) and (3) representing the two readings of (1).

(1) Everybody didn't pay attention.

(2)

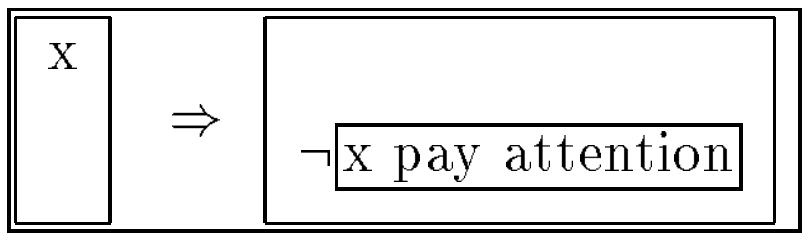

(3)

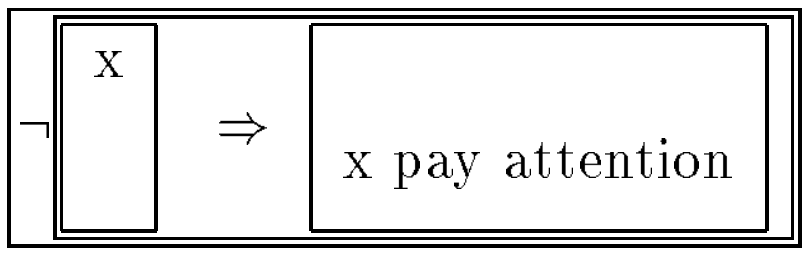

The following representations make the distinction between structure and content more explicit. The subordination relation $\leq$ is read from bottom to top.

(4)

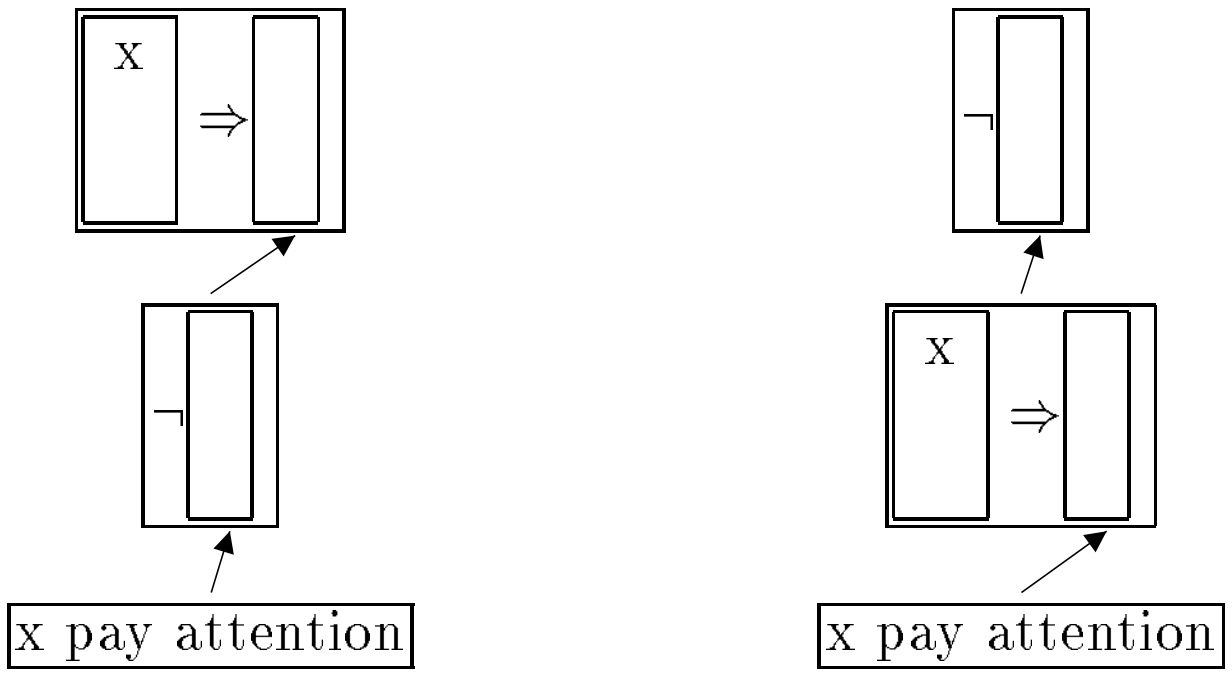

Having achieved this separation we are able to represent the structure that is common to both, (2) and (3), by (5).

(5)

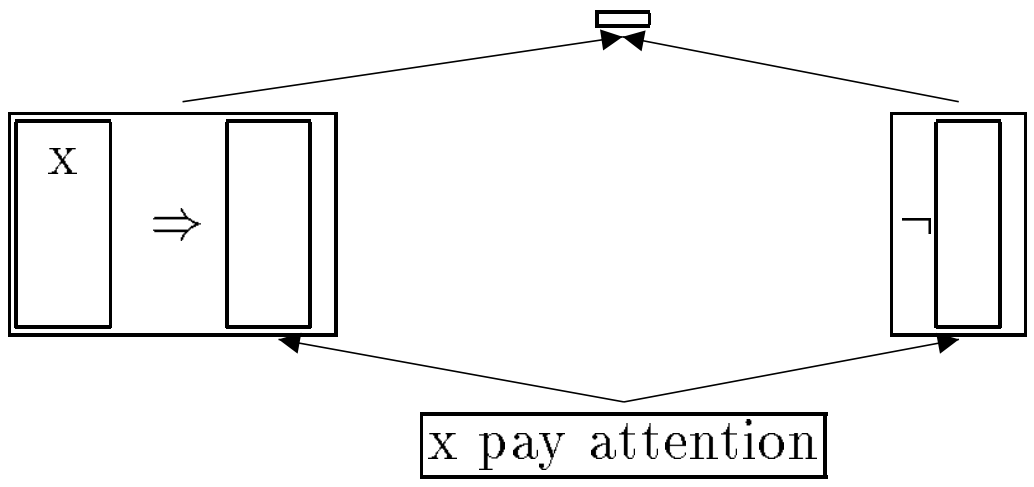

(5) is already the UDRS that represents (1) with scope relationships left unresolved.

To be more precise, we express the structural information by a language with one predicate $\leq$ that relates individual constants $\mathbf{l}$, called *labels*. The constants are names for DRS's. $\leq$ corresponds to the subordination relation between them, i.e. the set of labels with $\leq$ is a upper semilattice with one-element (denoted by $\mathbf{l}_\top$). The constants are also used to position DRS-conditions at the right place in the hierarchy. This is done by writing $\mathbf{l}{:}\gamma$ for an occurrence of a DRS-condition $\gamma$ in a DRS named $\mathbf{l}$. Thus the DRS (2) of (1) is represented by (6).

(6)
$$
\begin{array}{ll}
\mathbf{l}_\top{:}\mathbf{l}_{11} \Rightarrow \mathbf{l}_{12} & \\
\mathbf{l}_{11}{:}\mathrm{x} & \mathbf{l}_2 \leq \mathbf{l}_{12} \\
\mathbf{l}_2{:}\neg \mathbf{l}_{21} & \mathbf{l}_3 \leq \mathbf{l}_{21} \\
\mathbf{l}_3{:}\text{x pay attention} &
\end{array}
$$

(6) lists only the subordination relations that are neither implicitly contained in the partial order nor determined by complex UDRS-conditions. This means that (6) implicitly contains the information that, e.g., $\mathbf{l}_{21} \leq \mathbf{l}_\top$, and also that $\mathbf{l}_{21} \leq \mathbf{l}_2$, $\mathbf{l}_{11} \leq \mathbf{l}_\top$ and $\mathbf{l}_{12} \leq \mathbf{l}_\top$. To increase readability we will abbreviate the information in (6) by structures like (7), in which we will annotate DRSs with labels only if needed.

(7)

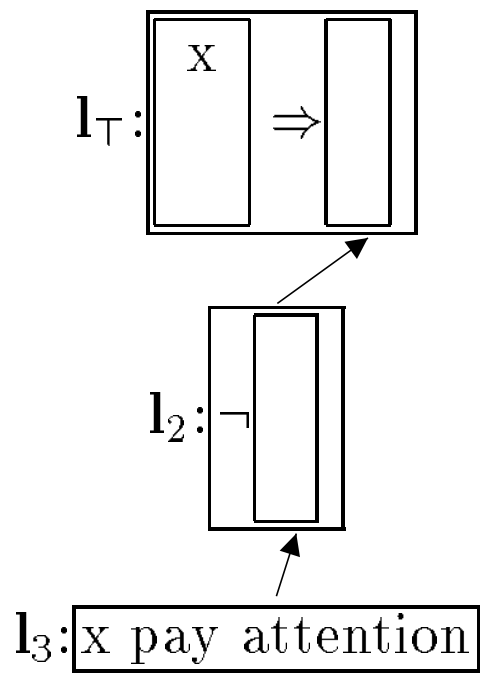

The underspecified representation of the two readings of (1) is given by (8),

(8)

$$\begin{array}{ll} \mathbf{l}_1{:}\mathbf{l}_{11}{\Rightarrow}\mathbf{l}_{12} & \mathbf{l}_1 \leq \mathbf{l}_\top \\ \mathbf{l}_{11}{:}\text{x} & \\ \mathbf{l}_2{:}\neg\mathbf{l}_{21} & \mathbf{l}_2 \leq \mathbf{l}_\top \\ \mathbf{l}_3{:}\text{x pay attention} & \mathbf{l}_3 \leq \mathbf{l}_{21}\ \mathbf{l}_3 \leq \mathbf{l}_{12} \end{array}$$

which is – if we do without annotating labels – the description of (5).

The construction of underspecified representations proceeds according to general principles such as, e.g., the conditions on Clause Boundedness, Scope of Indefinites and Proper Names.

(i) Let $\mathbf{l}_i$ be the label of a generalized quantifier and $\mathbf{l}_j$ the label associated with the top of the clause in which the quantified NP occurs. Then conditions of the form $\mathbf{l}_i \leq \mathbf{l}_j$ ensure that the scope of the quantifier is clause bounded.

(ii) The scope potential of an indefinite description, labelled $\mathbf{l}_i$, is not clause bounded. The fact that it may take arbitrarily wide scope, i.e. may get a more and more specific reading, is already captured by the condition introduced by (i). Of course it cannot exceed the toplevel DRS, i.e. $\mathbf{l}_i \leq \mathbf{l}_\top$.

(iii) Proper names, $\pi$, always end up in the top-level DRS, $\mathbf{l}_\top$. This is marked in the lexicon by $\mathbf{l}_\top{:}\pi$.

The following principle guarantees that no free variables are left in the representation irrespective of which disambiguation steps will be applied. To state the principle it is convenient to have the following definition. We define *scope* and *res* to be (partial) functions on the set of labels of a given UDRS $\mathcal{K}$. *scope* associates with each node the scope of this node, and *res* its restrictor.

(a) $scope(\mathbf{l}) = \mathbf{l}_1$ and $res(\mathbf{l}) = \mathbf{l}_1$, if $\{\mathbf{l}{:}\neg(\mathbf{l}_1)\}$ occurs in $\mathcal{K}$.

(b) $scope(\mathbf{l}) = \mathbf{l}_2$ and $res(\mathbf{l}) = \mathbf{l}_1$, if $\{\mathbf{l}{:}{\Rightarrow}(\mathbf{l}_1,\mathbf{l}_2)\}$ or $\{\mathbf{l}{:}\diamond(\mathbf{l}_1,\mathbf{l}_2)\}$ occurs in $\mathcal{K}$.

(c) $scope(\mathbf{l}) = \mathbf{l}$ and $res(\mathbf{l}) = \mathbf{l}$, if no condition of the forms mentioned in (a) and (b) and no condition of the form $\mathbf{l}{:}X$ occurs in $\mathcal{K}$, where X is a plural discourse referent.

According to condition (c) *scope* and *res* are only defined for labels that are introduced by non-ambiguous NPs, such as proper names, singular indefinites etc. They are not defined for plural NPs as long as their meaning is not disambiguated. We will come back to this shortly.

(iv) Suppose $\mathbf{l}_i$ is the label of the subDRS containing the verb and $\mathbf{l}_j$ the label of one of its arguments. Then the Closed Formula Principle ensures that the verb is in the scope of each of its arguments, simply by stipulating $\mathbf{l}_i \leq scope(\mathbf{l}_j)$.

The introduction called for a representation language that is able to directly express syntactic and semantic restrictions on quantifier scope. We give a simple example here and will fully implement a sophisticated syntactic theory of quantifier scope in the next sections.

(v) If the subject, labelled $\mathbf{l}_i$, must have scope over some other phrase, $\mathbf{l}_j$, then the condition $\mathbf{l}_j \leq scope(\mathbf{l}_i)$ is added.

It should be clear that the disambiguation of UDRSs is monotonic: If we add $\mathbf{l}_2 \leq \mathbf{l}_{12}$ to (8) we get a representation equivalent to (6). There is thus no need to restructure (parts of) a semantic representation if more information about scope restriction has become available. This process of enrichment is characteristic for the construction of UDRSs: Information from different sources (syntactic and semantic knowledge as well as knowledge about the world) may be incorporated in the structure by elaborating it in the sense just described. But let us now finish our introduction to UDRT by incorporating the analysis of plurals.

Plural NPs bear a high potential for creating ambiguities. As they can be understood either to denote a collection of individuals or to quantify over the members of that collection they give rise to the well-known collective/distributive ambiguity. But there are further possibilities to interpret sentences with plural NPs. (9.a) and (9.b) are examples of so-called generic and shared responsibility readings, respectively.

(9) a. The children in this city thrive.
b. The guys in 5b have been cheating on the exam again.

These readings differ from the distributive reading in that they can be accepted as true even if not all members of the set denoted by the subject NP are in the extension of the predicate expressed by the VP. To see that they differ from the collective reading for a similar reason consider (10).

(10) The girls gathered in the garden.

(10) has only a collective reading. It is true only if each of the girls goes to the garden with the intention to meet the others. The means that a predicate P is true of a group X, if every member of X contributes in some way or other to the fact that P is true of X. In (10) the contribution is the same for each girl and consists of having the property of intentionally going to the garden to meet the others. The generic and shared responsibility readings of (9) differ from the collective readings because they can be accepted as true even if not all members of the set denoted by the subject NP are in the extension of predicates that stand in such a relation to the VP. To specify the relevant relations is task of the lexical theory (and the specification of world knowledge). The task of UDRT is to provide an underspecified representation for all of these readings (for details see [Reyle 94]).

Collective and distributive uses of a verb $\gamma$ are determined by the type of discourse referents $\gamma$ takes. The UDRS in (12), for example, represents the collective reading of (11). (Discourse referents of type group are represented by capital letters.)

(11) The lawyers hired a secretary.

(12)

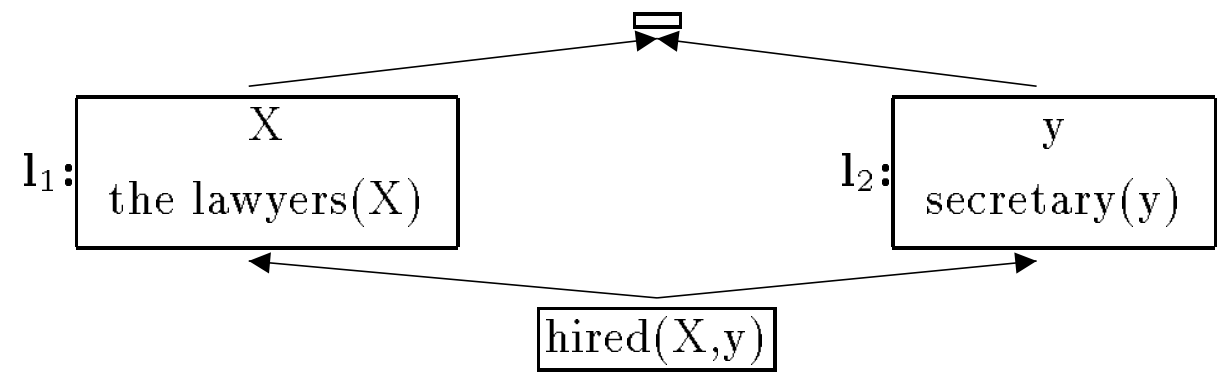

And its distributive reading is given in (13), where the quantification over the individual lawyers introduces a discourse referent, x, of type individual.

(13)

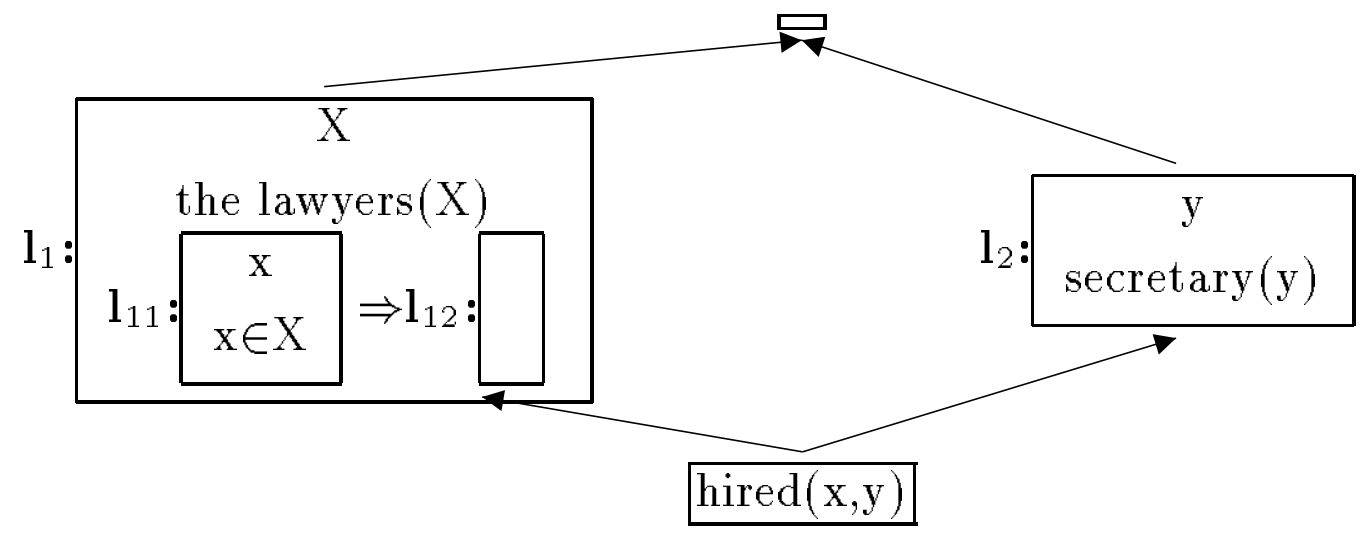

Let us note that although (12) is not ambiguous any more the choice of the distributive reading (13) for (11) leaves leeway for a further ambiguity. This ambiguity is due to the fact that the node representing the subject NP has been turned into a scope-bearing node by applying distribution to **the lawyers(X)**. Thus the indefinite can be interpreted as being within the scope of the distribution, or not. In (12) the NP-node is not scope-bearing, and, therefore, the UDRS is equivalent to the DRS that results by taking the union/merge of all subDRSs of (12).

In order to come to a representation that is underspecified with respect to the choice of possible readings of (11), we need to mark nodes to which a distribution might be applied as *potentially* scope bearing. As every label comes with *res* and *scope* we define:

(14) **(i)** $\mathbf{l}$ is *scope bearing* if $scope(\mathbf{l}) \neq \mathbf{l}$.

**(ii)** $\mathbf{l}$ is *not scope bearing* if $scope(\mathbf{l}) = res(\mathbf{l}) = \mathbf{l}$.

**(iii)** otherwise $\mathbf{l}$ is *potentially scope bearing*.

This definition will be used to give the underspecified representation of (11). It has exactly the shape of (12), but the argument DRS $\mathbf{l}_1$ is still marked as potentially scope bearing. Therefore *scope* and *res* are not yet defined for $\mathbf{l}_1$. Their values will only be given in the disambiguated representations. But there is a further complication with (12). It has to do with the discourse referents occurring as arguments of the verb: as long as we do not know if a distributive reading will be chosen, we do not want to use the referent standing for a group to occupy this position. Nevertheless we want to indicate the map between NP meanings and argument slots of the verb. This map is easily defined as follows.

Note that the construction of UDRSs guarantees a one-one correspondence between labels and discourse referents: if a UDRS $\mathcal{K}$ contains **l**:x and **l**:y then x = y. Let *dref* be the (partial) function associating with (the label of) a partial DRS its distinguished discourse referent, i.e. *dref*(**l**) = x if **l**:x is part of $\mathcal{K}$. Then *dref*(*res*(**l**)) gives us the following.[6]

**(i)** *dref*(*res*(**l**)) = x if **l** is introduced by a quantified NP, or by a plural NP which is interpreted distributively, i.e. **l**: 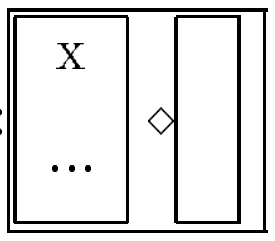

is part of $\mathcal{K}$.

**(ii)** *dref*(*res*(**l**)) = x if **l** is introduced by an indefinite singular NP, i.e. **l**: 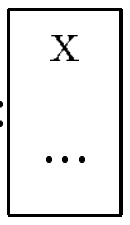

is part of $\mathcal{K}$.

**(iii)** *dref*(*res*(**l**)) = X if **l** is introduced by a plural NP which is interpreted collectively, i.e. **l**: 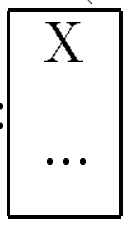

is part of $\mathcal{K}$.

If we use the term *dref*(*res*($\mathbf{l}_1$)) to specify which NP occupies which argument slot of the verb, we may replace (12) by (15) without changing its meaning.

(15)

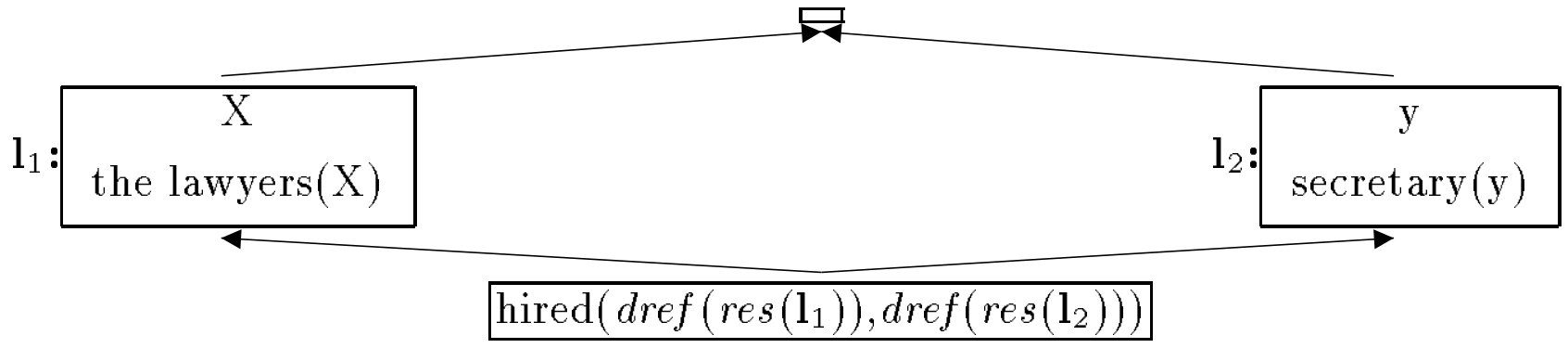

That (12) is only a notational variant of (15) is due to the fact that *res* is defined for $\mathbf{l}_1$ in (12). But if we assume that *res* is not defined for $\mathbf{l}_1$, then (15) represents the meaning of (11) in an underspecified way. Adding the information that *res*($\mathbf{l}_1$) = $\mathbf{l}_1$ disambiguates (15) and yields the collective reading. And adding the additional conditions $\mathbf{l}_1$:$\mathbf{l}_{11} \Rightarrow \mathbf{l}_{12}$, $\mathbf{l}_{11}$:x and $\mathbf{l}_{11}$:x∈X to (15) plus the information that *res*($\mathbf{l}_1$) = $\mathbf{l}_{11}$ results in disambiguating (15) to the distributive reading in (13).

In a similar way the choice of a generic or shared responsibility reading can be dealt with. Both introduce a quantificational structure. The generic reading, e.g., for (9.a) may be represented by (16), in which GEN denotes the generic quantifier.

[6] Distinguished discourse referents are referents introduced by indefinite (singular or plural) NPs, or by quantifiers. We will not consider cases of n-ary quantification in this paper.

(16)

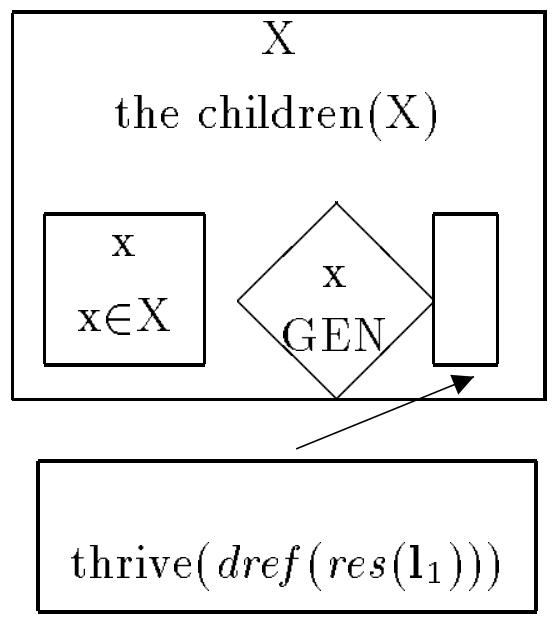

This method applies also to cumulative readings which are available when a verb is accompanied with two plural NPs, as in (17).

(17) Three breweries supplied five inns.

Under the cumulative reading (17) can be accepted as true if for each of the three breweries there is at least one inn the brewery supplies, and each inn is supplied by at least one brewery. The choice of this reading brings about the transition from (18) – which we may momentarily consider as the underspecified representation of (17) – to (19).

(18)

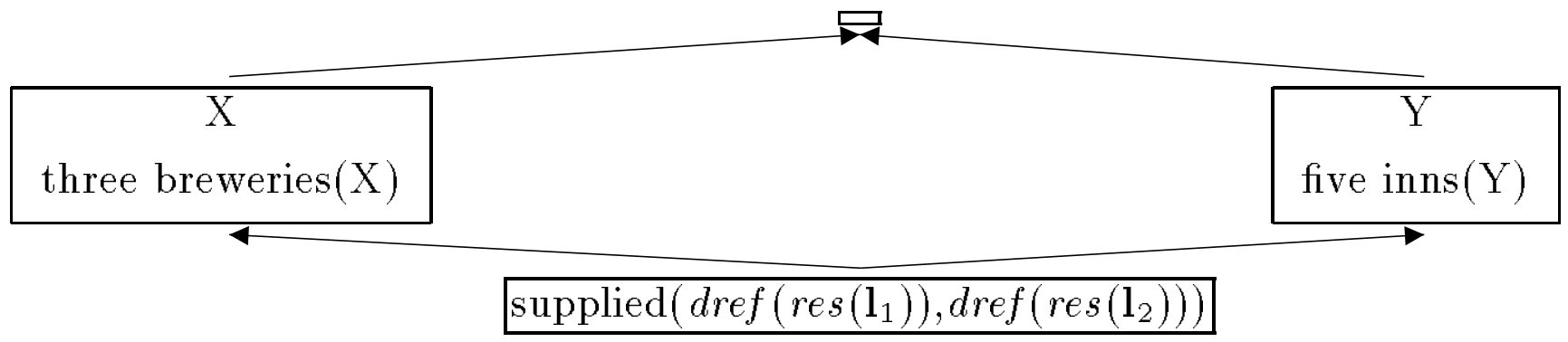

(19)

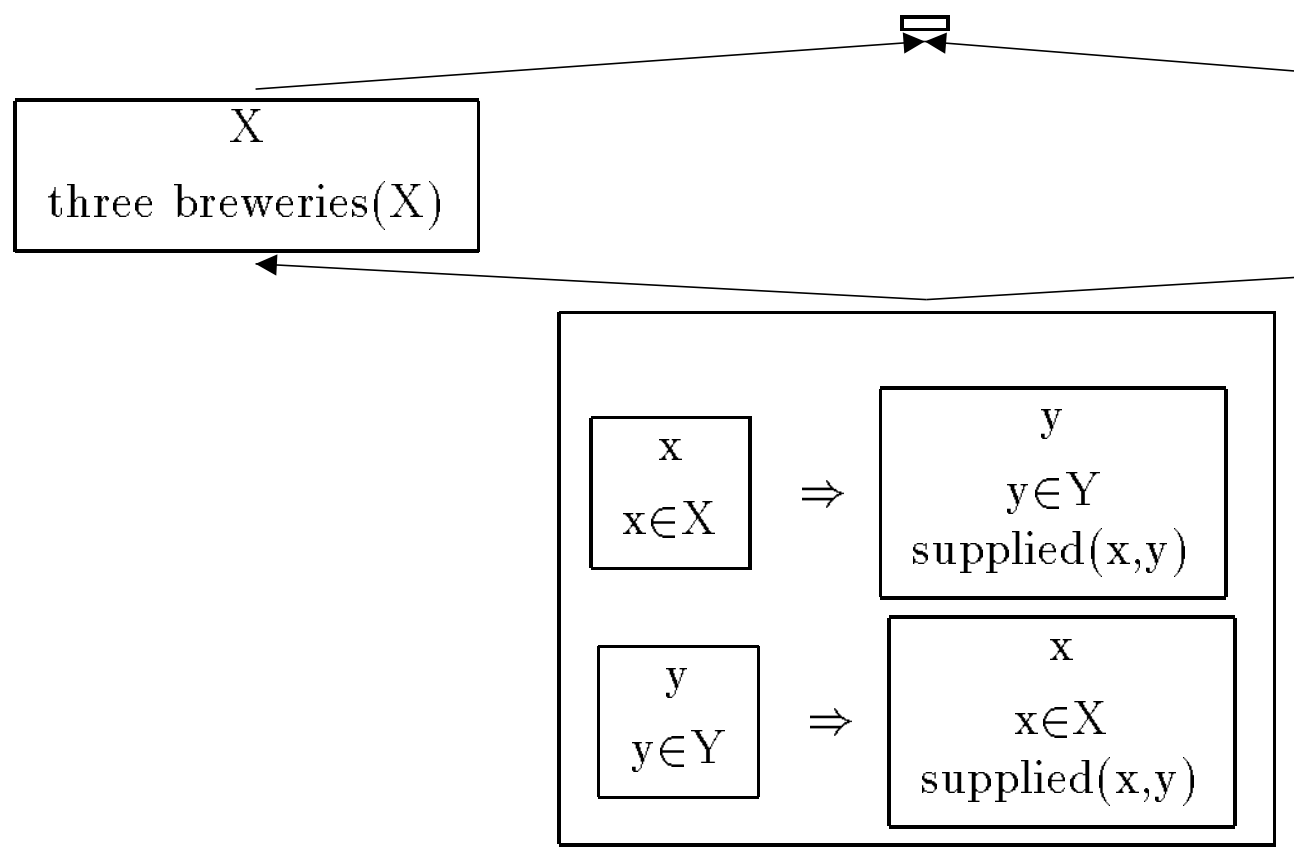

## 3 Syntactic Constraints on Quantifier Scoping

Work by Frey and Tappe (see [Frey] and [Frey/Tappe]) has shown that in German the relation between the actual positions occupied by the quantificational argument phrases of the verb and their traces are instrumental in determining the possible scope relations between the arguments.[7] In (20) for example *mindestens einen Bewerber* may have wide scope over *fast jedem Mitarbeiter* because the former NP c-commands the latter; and *fast jedem Mitarbeiter* may have wide scope over *mindestens einen Bewerber* because it c-commands the trace of *mindestens einen Bewerber*.

(20) Mindestens einen Bewerber habe ich fast jedem Mitarbeiter vorgestellt.

(21) [ [Mind. einen Bewerber]$_1$ habe [ ich f. jed. Mitarbeiter t$_1$ vorgestellt]]

If on the other hand *mindestens einen Bewerber* is not moved into the "Vorfeld", then it cannot take wide scope over any of the other NP's. This is shown by the non-ambiguous sentence

(22) Ich habe fast jedem Mitarbeiter mindestens einen Bewerber vorgestellt.

Frey and Tappe assume that all the argument phrases of German verbs (including their subjects) are dominated by the verb's maximal projection, $V^{max}$. If the arguments have been moved from their so-called *base position* they leave traces that are coindexed with the moved arguments (compare t$_1$ in (21)).[8] The movements that are relevant for the determination of scope ambiguities are, however, restricted to those occurring within – what is called – the local domain of the moved NP. This is exemplified by the non-ambiguity of examples like *Fast jeden Besucher meinte mindestens einer habe Maria gekannt*, in which the local domain of the NP *Fast jeden Besucher* is – roughly speaking – the complement structure of the matrix verb. In GB-terms the precise definition is as follows:

The *local domain* of an expression $\alpha$ is defined as the minimal complete functional complex, containing all licensing elements of $\alpha$ where a complete functional complex is defined as the minimal maximal projection in which all $\Theta$-roles are realized.

Given the notion of local domain we are able to state Frey's scope principle.[9]

(23) **Syntactic Scope Principle**
Suppose $L_\alpha$ is the local domain of an expression $\alpha$. Then $\alpha$ *may have scope over* an expression $\beta$ if either $\alpha$ or one of its traces c-commands $\beta$ itself or one of $\beta$'s traces in $L_\alpha$.

There is – to our knowledge – no syntactic theory that restricts scope relations between quantifiers and distributively interpreted plural NPs. We are, however, convinced that distributively interpreted plural NPs behave like genuine quantifers in all respects that are relevant for scoping relations. This assumption is supported by examples (24) – (25).

(24) Mindestens ein Mann glaubte, daß die Kinder Klingelputz gemacht haben.

[7] By "quantificational" argument phrase we understand a real generalized quantifier. This means that indefinites are not quantificational and thus not subject to the restrictions discussed.

[8] The basic order of verbal arguments can be identified in neutral intonation contexts which itself can be determined by focus projection tests). See, e.g. [Hoehle].

[9] We mentioned earlier that this principle may be applied also to adjuncts ([Frey/Tappe]). For reasons of space we cannot even touch the matter in this paper.

(25) a. weil der Mann mindestens einer Frau die Gemälde gezeigt hat.
b. weil der Mann die Gemälde mindestens einer Frau gezeigt hat.

The scope of the distributively interpreted plural in (24) is clause bounded, i.e. the distribution does not take scope over the quantifier *mindestens ein Mann*. Similarly, the relative scope between quantified NPs and distributively interpreted plural arguments in (25.a) and (25.b) corresponds to the Syntactic Scope Principle: in (25.a) the distribution does not take scope over the c-commanding NP *mindestens eine Frau*, whereas (25.b) has – besides the reading of (25.a) – a reading in which the scrambled plural NP *die Gemälde* may take scope over the quantified NP *mindestens einer Frau*.[10]

# 4 UDRS Construction in HPSG

In the following we will design a syntax-semantics interface for the construction of UDRSes in HPSG, focussing on the underspecified representation of scope and plural. To overcome the problems we pointed out in the introduction we will modify the standard HPSG framework in [Pollard/Sag] in several respects: The structure of the CONTENT attribute as well as the Semantics Principle will be changed substantially, since the construction of (U)DRSes allows for inherently different information structures and processing mechanisms. We replace the disambiguating mechanism for quantifier scope as it is realized in HPSG by use of a Cooper store by a principle-based description of scoping conditions based on partially ordered structures. Instead, using underspecified representations for scope ambiguities, we *monotonically* add scoping restrictions *only* when there is evidence for non-ambiguous scoping relations. For the determination of syntactic conditions on quantifier scope, in order to reconstruct Frey's scope principle in the HPSG formalism, besides the valence features SUBJ, COMPS, etc. proposed in [Pollard/Sag], we will introduce a head feature SUBCAT, which allows us to cope with scrambling and scope.[11] Moreover, contrary to [Pollard/Sag] we will assume binary branching structures.

## 4.1 The Representation of UDRSes in HPSG

A UDRS is represented as a complex feature structure UDRS, to replace the former CONTENT. It has the attributes SUBORD, CONDS and LS. SUBORD contains the information about the partial order of labels. It is defined as a set of subordination restrictions defined over labels. CONDS consists of a set of conditions $\gamma_i$, partial DRSes, which are associated with these labels by coindexation. The form of the labelled conditions is determined by the lexical entries. The attribute LS defines the distinguished labels, which indicate the upper and lower bounds for a partial DRS within the semilattice. Roughly speaking, one can say that, if $\mathbf{l}$ is the label of a DRS then $\mathbf{l}_{max}$ corresponds to $\mathbf{l}$ and $\mathbf{l}_{min}$ to *scope*($\mathbf{l}$).

[10] We do not present any examples containing two plural NPs with distributive interpretation, because the relative scope of two such NPs is hard to test due to the two *universal* quantifiers involved in their representation. Yet this does not contradict our assumption that distributive plural NPs obey the Syntactic Scope Principle.

[11] The use of a head feature SUBCAT has been proposed in [Frank] for the analysis of verb second in German.

(26)
$$\begin{bmatrix} \text{LOC} & \begin{bmatrix} \text{CAT} & cat \\ \text{UDRS} & \begin{bmatrix} \text{LS} & \begin{bmatrix} \text{L-MAX} & \mathbf{l}_{max} \\ \text{L-MIN} & \mathbf{l}_{min} \end{bmatrix} \\ \text{SUBORD} & \{\mathbf{l} \leq \mathbf{l}', ...\} \\ \text{CONDS} & \{\gamma_i, ...\} \end{bmatrix} \end{bmatrix} \end{bmatrix}$$

## 4.2 The Semantics Principle

The construction of UDRSes will be performed by clauses of the Semantics Principle: In (27), clause (I) of the Semantics Principle defines the inheritance of the partial DRSes defined in the CONDS attributes of the daughters to the CONDS value of the phrase. Contrary to the Semantics Principle of [Pollard/Sag] the semantic conditions are always inherited from *both* daughters, and therefore project to the uppermost sentential level. Furthermore, Clause (I) applies to *head-comp-* and *head-adj-structures* in exactly the same way.
Clause (II) of the Semantics Principle defines the inheritance of subordination restrictions: The subordination restrictions of the phrase are defined by the union of the SUBORD values of the daughters.[12]
Clause (III) of the Semantics Principle states the distinguished labels LS of the phrase to be identical to the distinguished labels of the HEAD-daughter. The role of the distinguished labels for UDRS construction will be discussed shortly.

**Semantics Principle:**[13]
**(I) Inheritance of UDRS-Conditions**
**(II) Inheritance of subordination restrictions**
**(III) Projection of the distinguished labels**[14]

(27)
$$\begin{bmatrix} \text{SYM} \mid \text{LOC} \mid \text{UDRS} & \begin{bmatrix} \text{LS} & \boxed{5} \\ \text{SUBORD} & \ldots \cup \boxed{3} \cup \boxed{4} \\ \text{CONDS} & \boxed{1} \cup \boxed{2} \end{bmatrix} \\ \textit{head-comp-struc} & \end{bmatrix}$$

C-DTR:
$$\begin{bmatrix} \text{SYM} \mid \text{LOC} \mid \text{UDRS} & \begin{bmatrix} \text{SUBORD} & \boxed{4} \\ \text{CONDS} & \boxed{2} \end{bmatrix} \end{bmatrix}$$

H-DTR:
$$\begin{bmatrix} \text{SYM} \mid \text{LOC} \mid \text{UDRS} & \begin{bmatrix} \text{LS} & \boxed{5} \\ \text{SUBORD} & \boxed{3} \\ \text{CONDS} & \boxed{1} \end{bmatrix} \end{bmatrix}$$

---

[12]The dots indicate that further subordination restrictions will be unioned to the phrase's SUBORD value. These will be defined below by the clauses (IV), (V) and (VI) of the Semantics Principle, which deal with binding of variables and (underspecified) quantifier scope.

[13]In the following the Semantics Principle will only be given for *head-comp-structures*. For *head-subj-* and *head-adj-structures* corresponding clauses have to be stated. For *head-filler-structures* we only have to state the inheritance of CONDS, SUBORD, and LS along the head projection, i.e. from the HEAD-DTR.

[14]Functional categories (determiners and complementizers) must be defined to inherit the distinguished labels of their complement. Thus the distinguished labels are projected along the *extended*, i.e. functional head projection in the spirit of [Grimshaw]. See the type definition *func-cat* for functional categories in Section 5.1.

The main task in constructing UDRSs consists in appropriately relating the labels of the partial DRSes that are to be combined. This is performed by the association of partial DRS conditions with distinguished labels in the lexicon entries on the one hand and by conditions governing the projection of the distinguished labels on the other. The role of the distinguished labels is most transparent with verbs and quantifiers.

In the lexicon entry of the verb, (28), the partial DRS is defined in the attribute CONDS as a relation holding between discourse referents. We have argued in Section 2 that we need functional terms of the form *dref*(*res*($\mathbf{l}_{max}$)) to occupy the argument positions, where $\mathbf{l}_{max}$ is the maximal label of the DP. We will implement this idea by combining the unary functions *dref* and *res* defined in Section 2 into a binary function *dref_res*, which takes as its first argument the entire content, i.e. the UDRS of the respective verb argument and returns the appropriate discourse referent to fill the argument position of the verb. If the NP argument is a plural, then the value of *dref_res* depends on the particular interpretation of the plural NP to be chosen. The remaining arguments of the function *dref_res* and its application for the account of plural underspecification will be discussed in detail in Section 5.2.
The partial DRS of the verb thus specified is associated with an identifying label **l**. It is defined as the minimal distinguished label of the verbal projection by coindexation with L-MIN.

(28)
$$\left[\text{LOC}\left[\begin{array}{ll}\text{CAT} \mid \text{HEAD} \mid \text{SUBCAT} < \text{DP}\begin{bmatrix}\text{CASE } \textit{nom}\\ \text{UDRS } \boxed{1}\end{bmatrix}, \text{DP}\begin{bmatrix}\text{CASE } \textit{dat}\\ \text{UDRS } \boxed{2}\end{bmatrix}, \text{DP}\begin{bmatrix}\text{CASE } \textit{acc}\\ \text{UDRS } \boxed{3}\end{bmatrix} > & \\ \text{UDRS}\begin{bmatrix}\text{LS}\left[\text{L-MIN } \boxed{1}\right]\\ \text{SUBORD } \{\}\\ \text{CONDS}\left\{\begin{bmatrix}\text{LABEL } \boxed{1}\\ \text{REL } \textit{vorstellen}\\ \text{ARG1 } \textit{dref_res}(\boxed{1}, \textit{Cond1}, \textit{Dref1})\\ \text{ARG2 } \textit{dref_res}(\boxed{2}, \textit{Cond2}, \textit{Dref2})\\ \text{ARG3 } \textit{dref_res}(\boxed{3}, \textit{Cond3}, \textit{Dref3})\end{bmatrix}\right\}\end{bmatrix} & \end{array}\right]\right]$$

It was mentioned in Section 2 that the partial structure of the verb has to be (weakly) subordinate to the scope of all the partial DRSes that introduce the discourse markers corresponding to the verb's arguments. This guarantees that all occurrences of discourse markers are properly bound by some superordinated DRS. The constraint is realized by clause (IV) of the Semantics Principle, the Closed Formula Principle. It guarantees that the label associated with the verb, which is identified with the distinguished minimal label of the sentential projection, is subordinated to the minimal label, or lower bound of each of the verb's arguments. Recall that with quantified arguments the predicate of the verb must be subordinate to the nuclear scope of the quantifier. We will see shortly that it is in fact the nuclear scope of the quantified structure that will be accessed by the distinguished minimal label of the quantified NP.

The Closed Formula Principle, shown in (29), states that in every (non-functional[15]) *head-complement-structure* a further subordination restriction is unioned to the phrase's SUBORD value, which subordinates the minimal label of the head – here the minimal label associated with the verb – to the minimal label of its actual complement, which in case of a quantified argument identifies the nuclear scope. Recall that due to clause (III) of the Semantics Principle it is guaranteed that in binary branching structures the minimal label of the verb as well as the distinguished minimal labels of the verb arguments are available all along their respective extended projections.

**Semantics Principle:**
**(I) & (II) & (III) & (IV) Closed Formula Principle**

(29)

$$\begin{bmatrix} \ldots \mid \text{UDRS} \begin{bmatrix} \text{LS} & \boxed{5} \\ \text{SUBORD} & \ldots \cup \{ \boxed{\mathbf{l}_{min}} \geq \boxed{\mathbf{l}_{verb}} \} \cup \boxed{3} \cup \boxed{4} \\ \text{CONDS} & \boxed{1} \cup \boxed{2} \end{bmatrix} \\ _{head-comp-struc} \end{bmatrix}$$

$$\text{C-DTR} \qquad\qquad\qquad\qquad \text{H-DTR}$$

$$\begin{bmatrix} \ldots \mid \text{UDRS} \begin{bmatrix} \text{LS} & \begin{bmatrix} \text{L-MIN} & \boxed{\mathbf{l}_{min}} \end{bmatrix} \\ \text{SUBORD} & \boxed{4} \\ \text{CONDS} & \boxed{2} \end{bmatrix} \end{bmatrix} \qquad \begin{bmatrix} \ldots \mid \text{UDRS} \begin{bmatrix} \text{LS} & \boxed{5} \begin{bmatrix} \text{L-MIN} & \boxed{\mathbf{l}_{verb}} \end{bmatrix} \\ \text{SUBORD} & \boxed{3} \\ \text{CONDS} & \boxed{1} \end{bmatrix} \end{bmatrix}$$

We could have chosen to encode the subordination restrictions governed by the Closed Formula Principle directly in the verb's lexicon entry instead of applying the principle during the syntactic analysis of the verbal projection. The SUBORD value would then be defined by reference to the verb's minimal label and the minimal labels of the subcategorized arguments in the verb's lexicon entry given above. This lexically determined mechanism would, however, not extend to a treatment of adjuncts, which must be syntax-driven.

Another motivation for the use of distinguished minimal and maximal labels comes from the representation of generalized quantifiers. Generalized quantifiers, as illustrated in (30) introduce two new labels identifying the partial DRSes of restrictor and scope. The quantificational relation holding between them is stated in terms of the relation attribute. Below we give the lexicon entry for *fast jeder*. A new discourse referent is introduced in the restrictor DRS, labelled $\mathbf{l}_{11}$, which is identified with the label of the subcategorized NP. The feature SUBORD defines the labels of restrictor and scope to be subordinate to the label $\mathbf{l}_1$ which identifies the entire condition. The label $\mathbf{l}_1$ is defined as the upper bound, or distinguished maximal label of the quantificational structure, whereas the lower bound, or distinguished minimal label is given by the label of the nuclear scope, $\mathbf{l}_{12}$. As $\mathbf{l}_1 \neq \mathbf{l}_{12}$ we mark the generalized quantifier as *scope bearing* (see (14.i)).

[15] Again, the exceptional case of functional heads, which are not to be subordinated to their functional complement, must be captured by additional constraints. In the framework of functional HPSG ([Netter], [Frank]), this can be done straightforwardly.

$$(30)\quad \left[\text{LOC}\left[\begin{array}{ll}\text{CAT} & \left[\begin{array}{ll}\text{HEAD} & \textit{quant}\\ \text{COMPS} & <\text{NP}\left[\text{LABEL}\ \boxed{\mathbf{l}_{11}}\right]>\end{array}\right]\\ \text{UDRS} & \left[\begin{array}{ll}\text{LS} & \left[\begin{array}{ll}\text{L-MAX} & \boxed{\mathbf{l}_1}\\ \text{L-MIN} & \boxed{\mathbf{l}_{12}}\end{array}\right]\\ \text{SUBORD} & \{\boxed{\mathbf{l}_1} > \boxed{\mathbf{l}_{11}}, \boxed{\mathbf{l}_1} > \boxed{\mathbf{l}_{12}}\}\\ \text{CONDS} & \left\{\left[\begin{array}{ll}\text{LABEL} & \boxed{\mathbf{l}_1}\\ \text{REL} & \textit{fast jeder}\\ \text{RES} & \boxed{\mathbf{l}_{11}}\\ \text{SCOPE} & \boxed{\mathbf{l}_{12}}\end{array}\right], \left[\begin{array}{ll}\text{LABEL} & \boxed{\mathbf{l}_{11}}\\ \text{DREF} & x\end{array}\right]\right\}\end{array}\right]\end{array}\right]\right]$$

Since we adopt a DP analysis, subordination constraints and discourse referents are introduced in the entries of the determiners or quantifiers, while the entries for nouns in (31) and (32) are almost trivial.

$$(31)\quad \left[\text{LOC}\left[\text{UDRS}\left[\begin{array}{ll}\text{SUBORD} & \{\}\\ \text{CONDS} & \left\{\left[\begin{array}{ll}\text{LABEL} & \mathbf{l}\\ \text{REL} & \textit{Mitarbeiter}\end{array}\right]\right\}\end{array}\right]\right]\right]$$

Proper names and indexicals will always be associated with the top label $\mathbf{l}_\top$, so that they automatically end up in the main DRS.

$$(32)\quad \left[\text{LOC}\left[\text{UDRS}\left[\begin{array}{ll}\text{SUBORD} & \{\}\\ \text{CONDS} & \left\{\left[\begin{array}{ll}\text{LABEL} & \mathbf{l}_\top\\ \text{REL} & \textit{peter}\\ \text{DREF} & x\end{array}\right]\right\}\end{array}\right]\right]\right]$$

The entry for the indefinite singular determiner (33) only introduces a new discourse referent for individuals. The ensuing DP is marked as *not scope bearing* – in the sense of (14.ii) above – by the identity statement $\mathbf{l}_1 = \mathbf{l}_{12}$ for the minimal and maximal labels in the set of subordination restrictions.

$$(33)\quad \left[\text{LOC}\left[\begin{array}{ll}\text{CAT} & \left[\begin{array}{ll}\text{HEAD} & \left[\text{AGR} \mid \text{NUM}\ \ \textit{sg}\right]\\ \text{COMPS} & <\text{NP}\left[\text{LABEL}\ \boxed{\mathbf{l}_{12}}\right]>\end{array}\right]\\ \text{UDRS} & \left[\begin{array}{ll}\text{SUBORD} & \{\boxed{\mathbf{l}_1} = \boxed{\mathbf{l}_{12}}\}\\ \text{LS} & \left[\begin{array}{ll}\text{L-MAX} & \boxed{\mathbf{l}_1}\\ \text{L-MIN} & \boxed{\mathbf{l}_{12}}\end{array}\right]\\ \text{CONDS} & \left\{\left[\begin{array}{ll}\text{LABEL} & \boxed{\mathbf{l}_1}\\ \text{DREF} & x\end{array}\right]\right\}\end{array}\right]\end{array}\right]\right]$$

As we have seen in Section 2, in order to allow for an underspecified representation of plural NPs we have to define plural NPs as *potentially scope bearing*, in contradistinction to quantifiers – marked as *scope bearing* by non-identical values of minimal and maximal labels – and singular NPs – marked as *not scope bearing* by identifying minimal and maximal labels.

This can be achieved if we do not completely specify the relation between the minimal label $\mathbf{l}_{12}$ and the maximal label $\mathbf{l}_1$ in (34), but only require that $\mathbf{l}_{12}$ is weakly subordinate to $\mathbf{l}_1$. This weak subordination relation will be further restricted to either identity or strict subordination when more information is available from the semantic or pragmatic context that allows the ambiguity to be resolved. By monotonically adding further constraints a collective or quantificational (distributive or generic) reading of the plural NP may then be specified.[16] If a distributive reading is chosen, the minimal label $\mathbf{l}_{12}$ will identify the nuclear scope of the quantified structure, and in the case of a collective reading the relation of (weak) subordination between minimal and maximal label will be reduced to identity. We will state this in detail in Section 5.2.

(34)

$$\left[\text{LOC}\begin{bmatrix}\text{CAT}\begin{bmatrix}\text{HEAD}\begin{bmatrix}\text{AGR}\mid\text{NUM}\ \ pl\end{bmatrix}\\ \text{COMPS}\ \ < \begin{bmatrix}\text{LABEL}\ \boxed{l_1}\end{bmatrix} >\end{bmatrix}\\ \text{UDRS}\begin{bmatrix}\text{SUBORD}\ \{\boxed{l_1}\geq\boxed{l_{12}}\}\\ \text{LS}\begin{bmatrix}\text{L-MAX}\ \boxed{l_1}\\ \text{L-MIN}\ \boxed{l_{12}}\end{bmatrix}\\ \text{CONDS}\ \left\{\begin{bmatrix}\text{LABEL}\ \boxed{l_1}\\ \text{DREF}\ \ X\end{bmatrix}\right\}\end{bmatrix}\end{bmatrix}\right]$$

Together with the structure of the lexical entries illustrated above, the clauses (I) – (IV) of the Semantics Principle given in (29) define the core mechanism for UDRS construction: The Semantics Principle defines the inheritance of the labelled DRS conditions, as well as the subordination restrictions between these labels, which give us the semilattice for the complete UDRS structure. The subordination restrictions are projected from the lexicon or get introduced monotonically, e.g., by the Closed Formula Principle, to ensure the correct binding of discourse referents. Further subordination restrictions will be added – monotonically – by the remaining clauses of the Semantics Principle, to be introduced in the next Section, which govern the interaction of quantificational scope and scrambling for real quantificational NPs as well as for quantificational structures induced by distributive or generic readings of plural NPs.

# 5 A Syntax–Semantics Interface for Scoping Principles: Quantifiers & Plural

Since the conditions on quantificational scope for generalized quantifiers and distributive readings of plural NPs are highly dependent on syntactic structure, the Semantics Principle will be supplemented by further clauses governing the interface between syntactic constraints and semantic representation. These principles will identify syntactic conditions for quantificational scope according to the theory of [Frey], and specify (partially) disambiguating semantic restrictions, respecting monotonicity.

[16] We are not in the position to discuss the factors that determine these constraints here.

## 5.1 Local Domain for Quantifier Scope

Recall what we said about the scope potential of indefinite NPs and genuine quantifiers. Whereas the former may take arbitrarily wide scope, the latter are allowed to take scope only over elements that appear in their local domain. The same restriction was argued to hold for distributive readings of plural NPs. We will implement this restriction by requiring that the maximal label of the plural NP must be subordinate to the distinguished label which identifies the upper bound of the local domain.

In [Frey] the local domain of an expression $\alpha$ is defined as the minimal complete functional complex containing the licensing elements of $\alpha$, where a complete functional complex is defined as the minimal maximal projection in which all $\Theta$-roles are realized (see (23)). By this definition, for finite sentences the local domain for a verb argument comes down to the local IP projection. In the functional HPSG grammar described in [Frank] this local domain corresponds to the functional projection of the finite VP in which all verb arguments have been saturated.[17] Thus the distinguished maximal label $\mathbf{l}_{max}$ which identifies the upper bound of the local domain for quantified verb arguments will be instantiated by the complementizer heading a finite sentence or by the finite verb in second position, which both are of type *func-cat*.

(35)

$$\begin{bmatrix} \text{LOC} \begin{bmatrix} \text{CAT} \begin{bmatrix} \text{COMPS} < \begin{bmatrix} \text{VFORM} & fin \\ \text{LS} & \boxed{1} \end{bmatrix} > \end{bmatrix} \\ \text{UDRS} \begin{bmatrix} \text{LS} & \boxed{1} \begin{bmatrix} \text{L-MAX} & \mathbf{l}_{max} \end{bmatrix} \end{bmatrix} \end{bmatrix} \\ {}_{func-cat} \end{bmatrix}$$

Again due to the projection of the distinguished labels LS by clause (III) of the Semantics Principle and the definition of functional categories, the upper bound for the local domain of quantifier scope, $\mathbf{l}_{max}$, is available throughout the extended projection, where clause (V) of the Semantics Principle, the Quantifier Scope Principle, applies.

The Quantifier Scope Principle (V) in (36) states that in a *head-comp-structure* where the complement is a generalized quantifier (type *quant*) or a potentially scope bearing plural NP (type *plural*) the SUBORD value of the phrase will contain the further condition that the complement's maximal label $\mathbf{l}_{quant}$ is subordinate to the label $\mathbf{l}_{max}$ which identifies the upper bound of the local domain, as defined above. The definition will be slightly revised for the case of distributive readings in Section 6.

[17] Similar definitions for the identification of local domains must be stated for other syntactic configurations, e.g. infinite complements, NPs, etc. We cannot go into these details here.

**Semantics Principle:**[18]
**(I) & (II) & (III) & (IV) &**
**(V) Quantifier Scope Principle** (provisional)

(36)

$$\left[\begin{array}{l} \ldots \mid \text{UDRS} \left[\begin{array}{l}\text{LS } \boxed{5} \\ \text{SUBORD } \ldots \cup \{\, \boxed{l_{max}} \geq \boxed{l_{quant}} \} \cup \{\, \boxed{l_{min}} \geq \boxed{l_{verb}} \} \cup \boxed{3} \cup \boxed{4} \\ \text{CONDS } \boxed{1} \cup \boxed{2}\end{array}\right] \\ \textit{head-comp-struc}\end{array}\right]$$

C-DTR:

$$\left[ \ldots \left[\begin{array}{l} \text{CAT} \mid \text{HEAD } \; \textit{quant} \vee \textit{plural} \\ \text{UDRS} \left[\begin{array}{l} \text{LS} \left[\begin{array}{l} \text{L-MAX } \boxed{l_{quant}} \\ \text{L-MIN } \boxed{l_{min}} \end{array}\right] \\ \text{SUBORD } \boxed{4} \\ \text{CONDS } \boxed{2} \end{array}\right] \end{array}\right] \right]$$

H-DTR:

$$\left[ \ldots \mid \text{UDRS} \left[\begin{array}{l} \text{LS } \boxed{5} \left[\begin{array}{l} \text{L-MAX } \boxed{l_{max}} \\ \text{L-MIN } \boxed{l_{verb}} \end{array}\right] \\ \text{SUBORD } \boxed{3} \\ \text{CONDS } \boxed{1} \end{array}\right] \right]$$

## 5.2 How to Cope with Scrambling and Scope

The Semantics Principle as it is defined up to now does not yet implement Frey's Scope principle (23). The UDRS built up for (22), with narrow scope of the direct object, would – incorrectly – come out the same as the one for the ambiguous sentence (20), namely an underspecified representation which leaves unresolved the relative scope of the quantified arguments. Therefore, the last clause of the Semantics Principle, the Complement Scope Principle (CSP) has to determine the relative quantificational scope according to Frey's Scope Principle, and monotonically add more and more scoping constraints to the subordination structure encoded by SUBORD. It is important to note that we start out with completely underspecified quantificational subordination restrictions in the verb's lexicon entry (SUBORD is the empty set) and monotonically introduce subordination restrictions only if there is evidence for non-ambiguous scoping relations.[19]

We already mentioned that our syntax–semantics interface is based on a functional HPSG-style grammar (see [Netter], [Frank]). As proposed in [Frank] for the analysis of verb second, besides the valence features SUBJ and COMPS, which are governed by the valence principle, we use the original SUBCAT list as a head feature, which will be projected by the Head Feature Principle. The order of the elements on SUBCAT will be lexically determined by the so-called 'basic' or 'normal order' of the arguments, which can be identified in neutral intonation contexts.[20] The instantiation of the valence features is defined by the Valence Instantiation Principle (VIP), here given for the lexical category types:

[18] In the actual implementation, a disjunction must be defined for *(potentially) scope bearing* vs *not scope bearing* arguments. Here, we only state the relevant disjunct which applies to *(potentially) scope bearing* arguments.

[19] We depart significantly, here, from our earlier account in [Frank/Reyle], where monotonicity was not ensured: Subordination constraints defining narrow scope of quantified arguments were stated in the verb's lexicon entry, and could be discarded from SUBORD, depending on the actual syntactic scoping configuration.

[20] See e.g. [Hoehle], [Haider], [Frey].

(37) $$\begin{bmatrix} \text{LOC} \mid \text{CAT} \begin{bmatrix} \text{HEAD} \mid \text{SUBCAT} & < \boxed{1} > \oplus \boxed{2} \\ \text{LEX} & + \\ \text{SUBJ} & < \boxed{1} > \\ \text{COMPS} & \boxed{2} \end{bmatrix} \\ _{lex-cat} \end{bmatrix}$$

Via the VIP principle, the lexically determined 'normal order' of verb arguments is carried over to the valence features. The valence principles, which replace the former subcat principle, are now defined to saturate the subcategorized arguments *in the order* determined by the SUBCAT list by requiring the actually processed argument to be identified with the *last* element on the respective valence list. Below we state the order sensitive valence principle for *head-comp-structures.*[21]

(38)

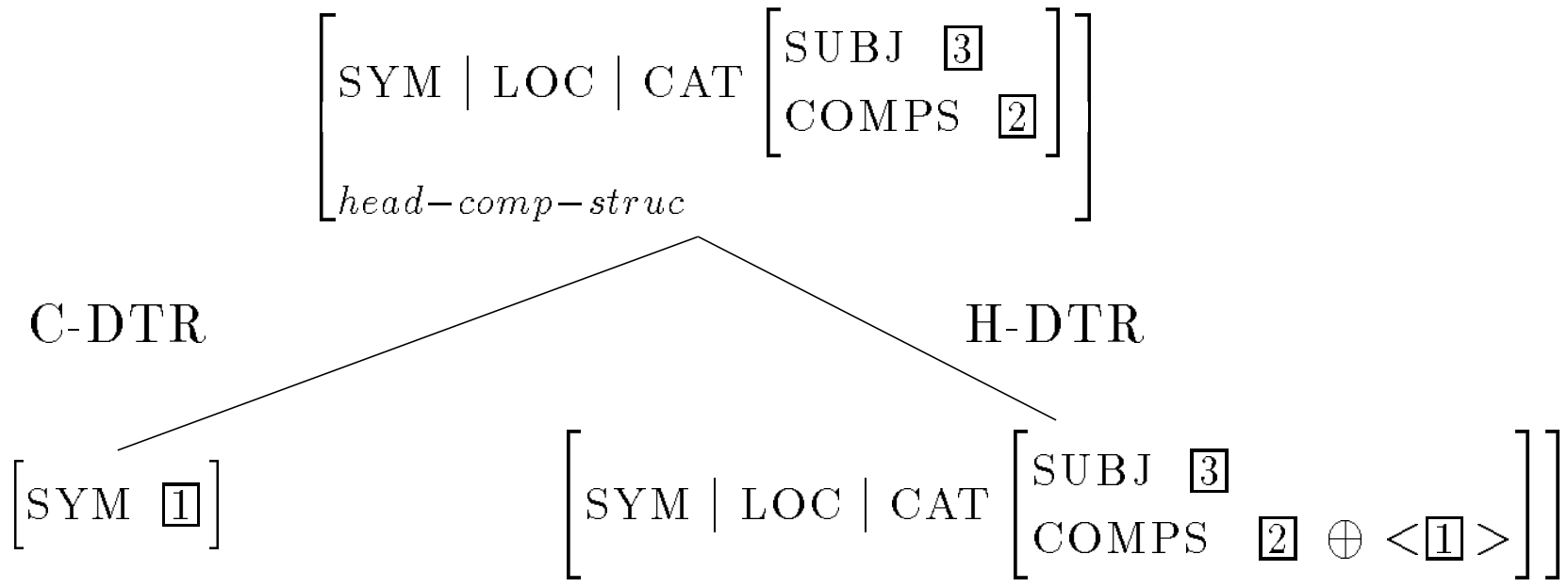

Verb arguments not showing up in the 'normal order' defined by the verb – being scrambled or topicalized – will be analyzed by a trace in the base position. Thus, the trace is analyzed by the valence principles as a non-overt complement or subject daughter, which gets identified with the last position of the respective valence feature, whereas – by the nonlocal feature mechanism – the overt antecedent will be introduced in a higher position as a filler daughter.

With these definitions at hand, we are in a position to state the Complement Scope Principle as a reconstruction of Frey's scope principle.
Due to the right-branching structure of the VP in German the precedence relations holding among the elements of the head's SUBCAT list correspond to the c-command relations that hold among the verb arguments if they appear in 'normal order'. Recall that we start out with underspecified scope relations. Thus the Complement Scope Principle has to introduce subordination restrictions *only* when there is evidence for *non-ambiguity*, i.e. when a quantifier takes necessarily wide scope over another quantifier. It suffices to consider one configuration, the *head-comp-* or *head-subj-structure* to make the point.
If the actually processed argument in a *head-comp-* or *head-subj-structure* is overtly realized, due to the ordering constraint of the valence principles, it is predicted to occur in its base position according to the 'normal order' encoded by SUBCAT. Given the correspondence of c-command relations between arguments and precedence relations between the elements on SUBCAT, if the actual argument is a scope bearing element, e.g. a quantifier, it is predicted to take scope over every quantified element that follows its position on the SUBCAT list – except for those which are to be found in the actual SLASH value. For, though in this case the actually processed argument takes undoubtly scope over the c-commanded trace of the slashed element, the overt antecedent will be realized in a higher position, from which it will c-command, in turn, the actually processed quantifier. Thus for those quantified elements which follow the quantified argument on SUBCAT, but which are contained in SLASH, we have ambiguous scoping relations and no scoping constraint must be added to SUBORD. For all those, however, which are *not* contained in SLASH, they are either realized in a c-commanded base position or are 'moved' into a c-commanded position, so wide scope of the actually processed quantifier must be fixed in the set of subordination restrictions.
The very same reasoning holds for the case in which the actually processed argument is a trace, corresponding to a quantificational antecedent. Again, given the correspondence of c-command and precedence on SUBCAT, the argument will take wide scope over every quantified element that follows it on the SUBCAT list – again except for those which are contained in SLASH and therefore will take scope over the actually processed trace from a higher position.
The reader may convince himself that it is not necessary to consider *head-filler-structures* for topicalized or scrambled arguments in order to determine the relative scope of quantified arguments.[22] We do only need to consider the 'base positions' of the arguments, i.e. *head-comp-* and *head-subj-structures.*

[21]The relative order of SUBJ and COMP daughters is ensured by requiring COMPS to be empty (saturated) in the valence principle for *head-subj-structures.*

Clause (VI) of the Semantics Principle, the Complement Scope Principle, can now be stated – provisionally – as follows: For every *head-comp-structure* with COMP-DTR *arg* or *head-subj-structure* with SUBJ-DTR *arg*, if *arg* is of type *quant* ∨ *plural*, then for every element $\alpha$ which follows *arg* on SUBCAT, which is of type *quant* ∨ *plural* and whose LOC-value is *not* contained in SLASH, the condition $\mathbf{l}_{min} \geq \mathbf{l}_{\alpha_{max}}$ is added to SUBORD, where $\mathbf{l}_{min}$ is the minimal label of *arg* and $\mathbf{l}_{\alpha_{max}}$ is the maximal label associated with each element $\alpha$.

**Semantics Principle:**
**(I) & (II) & (III) & (IV) & (V) &**
**(VI) Complement Scope Principle** (provisional)

$$\left[\begin{array}{l} ..\left[\begin{array}{ll} \text{LOC} \mid \text{UDRS} & \left[\begin{array}{l} \text{LS } \boxed{5} \\ \text{SUBORD } \boxed{10} \cup \{ \boxed{\mathbf{l}_{max}} \geq \boxed{\mathbf{l}_{quant}} \} \cup \{ \boxed{\mathbf{l}_{min}} \geq \boxed{\mathbf{l}_{verb}} \} \cup \boxed{3} \cup \boxed{4} \\ \text{CONDS } \boxed{1} \cup \boxed{2} \end{array}\right] \\ \text{NLOC} \mid \text{INH} \mid \text{SLASH } \boxed{9} & \end{array}\right] \\ _{head-comp-struc} \end{array}\right]$$

(39) C-DTR H-DTR

$$\left[..\boxed{7}\,\text{LOC}\left[\begin{array}{ll} \text{CAT} \mid \text{HEAD} & quant \vee plural \\ \text{UDRS} & \left[\begin{array}{l} \text{LS}\left[\begin{array}{ll} \text{L-MAX} & \boxed{\mathbf{l}_{quant}} \\ \text{L-MIN} & \boxed{\mathbf{l}_{min}} \end{array}\right] \\ \text{SUBORD } \boxed{4} \\ \text{CONDS } \boxed{2} \end{array}\right] \end{array}\right]\right] \quad \left[..\left[\begin{array}{ll} \text{CAT} \mid \text{HEAD} \mid \text{SUBCAT} & \boxed{6} \oplus \boxed{7} \oplus \boxed{8} \\ \text{UDRS} & \left[\begin{array}{l} \text{LS } \boxed{5} \left[\begin{array}{ll} \text{L-MAX} & \boxed{\mathbf{l}_{max}} \\ \text{L-MIN} & \boxed{\mathbf{l}_{verb}} \end{array}\right] \\ \text{SUBORD } \boxed{3} \\ \text{CONDS } \boxed{1} \end{array}\right] \end{array}\right]\right]$$

Condition: $\boxed{10}$ is the set of conditions of the form $\boxed{\mathbf{l}_{min}} \geq \boxed{\mathbf{l}_{\alpha_{max}}}$
where $\mathbf{l}_{\alpha_{max}}$ is the maximal label of every element $\alpha$ in $\boxed{8}$ such that $\alpha$ is of type *quant* ∨ *plural* and the LOC value of $\alpha$ is not contained in SLASH $\boxed{9}$[23]

[22] This also holds for so-called reconstruction cases, see [Frey].
[23] It is worth mentioning that the negative constraint about elements *not* contained in the SLASH value

# 6 Plural Disambiguation

We argued above that, due to the collective/distributive ambiguity of plural NPs their meanings has to be represented by *potentially scope bearing* partial DRSs. This was achieved by stating the minimal label of the plural NP to be weakly subordinated to its maximal label in (34). Together with the definition of the lexical entry of the verb as stated in (28), for example (11) we get the underspecified representation (40) (for ease also given in graph notation in (41)).

(40)
$$\left[\begin{array}{ll} \text{CAT} \mid \text{HEAD} \mid \text{SUBCAT} & < \text{DP}\left[\text{UDRS } \boxed{1}\right],\ \text{DP}\left[\text{UDRS } \boxed{2}\right] > \\ \text{UDRS} & \left[\begin{array}{ll} \text{SUBORD} & \{\, \mathbf{l}_\top \geq \boxed{l_1}, \mathbf{l}_\top \geq \boxed{l_2}, \boxed{l_1} \geq \boxed{l_{12}}, \boxed{l_{12}} \geq \boxed{l_3}, \boxed{l_2} \geq \boxed{l_3} \} \\ \text{CONDS} & \left\{ \left[\begin{array}{l} \text{LABEL } \boxed{l_1} \\ \text{REL } \textit{Rechts} \\ \qquad \textit{anwalt} \\ \text{DREF } X \end{array}\right], \left[\begin{array}{l} \text{LABEL } \boxed{l_2} \\ \text{REL } \textit{Sekr.} \\ \text{DREF } y \end{array}\right], \left[\begin{array}{l} \text{LABEL } \boxed{l_3} \\ \text{REL } \textit{einstellen} \\ \text{ARG1 } \textit{dref_res}(\boxed{1}, \textit{Cond1}, \textit{Dref1}) \\ \text{ARG2 } y \end{array}\right] \right\} \end{array}\right] \end{array}\right]$$

(41)

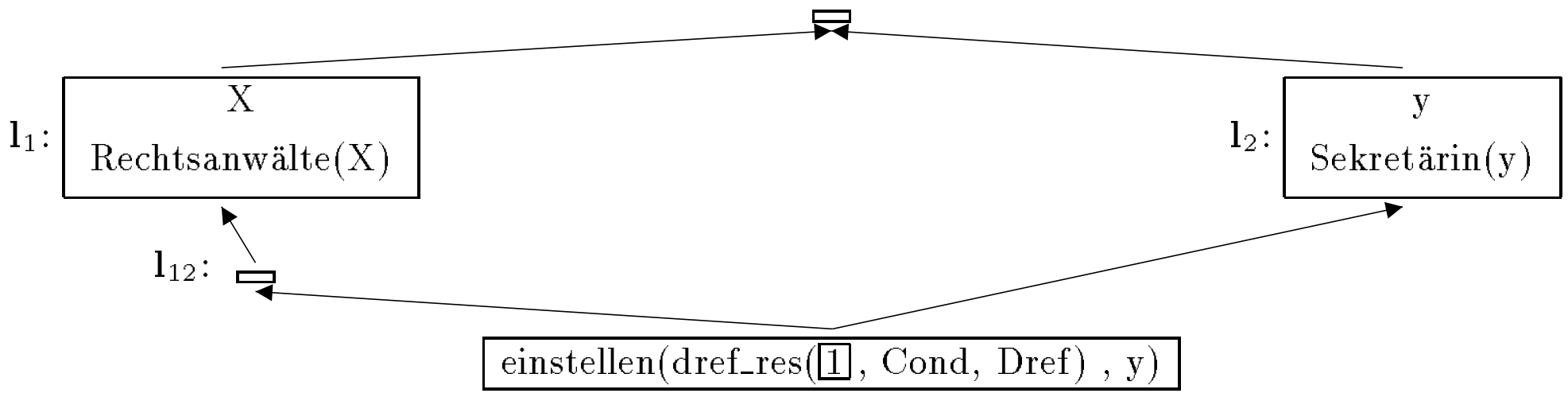

If context does not provide us with further, disambiguating information, (40) will be the final, underspecified representation for (11). While the value of the function *dref_res* is defined for the object NP and returns the individual type referent **y** introduced by it, it is undefined for the underspecified plural subject (compare the definition of *dref* (*res*(**l**)) in Section 2).

In a typed feature unification system, we cannot implement this requirement for an underspecified representation of plurals by using a type hierarchy or similar devices which come to mind straightforwardly. For it is *not* appropriate for the issue of underspecified representations to compute the set of disjunctive readings, which would ensue automatically if we took such an approach. Instead, the function *dref_res* will be implemented by using delaying techniques. The conditions which determine the delayed evaluation of the function *dref_res* are defined in its second argument *Cond*. As long as the variable *Cond* is not instantiated, the evaluation of *dref_res* will be blocked, i.e. *delayed*.[24]

---

can be implemented without using negation, which is too powerful for most feature structure formalisms. We calculate the difference list between the c-commanded elements $\boxed{8}$ and the list of elements in SLASH (which must be type-raised to *synsem*-objects $\boxed{\text{slash}}$): $\boxed{8} = \boxed{\text{diff-1}} \oplus \boxed{\text{slash}} \oplus \boxed{\text{diff-2}}$. The concatenation $\boxed{\text{diff-1}} \oplus \boxed{\text{diff-2}}$ then gives us the list of elements in $\boxed{8}$ which are not contained in SLASH.

[24] In the CUF system ([Doerre/Dorna]) delay statements are defined by the predicate *wait*. The specified argument positions of the delayed function are constrained to be instantiated in order for the delayed function to be evaluated.

In our example the delay statement for *dref_res* is defined as follows: *wait*(*dref_res*(*udrs, subord_info,* _)), where *udrs* and *subord_info* are the types of the value of UDRS and of a member of SUBORD respectively.

The three clauses of the function *dref_ref* in (42) distinguish between *not scope bearing*, *scope bearing* and *potentially scope bearing elements*, respectively. As we have seen above, in the HPSG semantics the distinction is defined in terms of the minimal and maximal labels, where the minimal label $\mathbf{l}_{min}$ identifies the *scope* of the maximal label $\mathbf{l}_{max}$ of a partial DRS.

Thus the first clause of (42), which takes as its first argument the UDRS value of a verb argument, as defined in (28), is only appropriate for non-quantificational singular NPs (33). The set of subordination conditions pertaining to the argument is constrained to contain a condition which identifies its minimal and maximal labels: $\mathbf{l}_1 = \mathbf{l}_{12}$.
The second clause applies in case the semantic structure of the verb argument contains a subordination restriction which characterizes the NP as *scope bearing*, as e.g. generalized quantifiers defined in (30). The values of the minimal and maximal labels are characterized as non-identical by a condition of strong subordination: $\mathbf{l}_1 > \mathbf{l}_{12}$.
If these clauses are applied successfully, by coindexation of the differentiating subordination restrictions with the second argument place of *dref_res*, the latter gets properly instantiated and the function is relieved from its delayed status. It returns the discourse referent which is defined in the argument's CONDS attribute for the maximal or restrictor's label, respectively, according to the definitions given in Section 2. For reasons to be discussed below this discourse referent in addition fills the third argument position.

(42)

$$dref_res\left(\begin{bmatrix} \text{LS}\begin{bmatrix}\text{L-MAX} & \boxed{\mathbf{l}_1}\\ \text{L-MIN} & \boxed{\mathbf{l}_{12}}\end{bmatrix}\\ \text{SUBORD}\{\ldots,\ \boxed{2}\left[\boxed{\mathbf{l}_1} = \boxed{\mathbf{l}_{12}}\right],\ \ldots\}\\ \text{CONDS}\left\{\ldots,\begin{bmatrix}\text{LABEL} & \boxed{\mathbf{l}_1}\\ \text{DREF} & \boxed{\text{x}}\end{bmatrix},\ \ldots\right\}\end{bmatrix},\ \boxed{2}\left[\boxed{\mathbf{l}_1} = \boxed{\mathbf{l}_{12}}\right],\boxed{\text{x}}\right) := \boxed{\text{x}}$$

$$dref_res\left(\begin{bmatrix} \text{LS}\begin{bmatrix}\text{L-MAX} & \boxed{\mathbf{l}_1}\\ \text{L-MIN} & \boxed{\mathbf{l}_{12}}\end{bmatrix}\\ \text{SUBORD}\{\ldots,\ \boxed{2}\left[\boxed{\mathbf{l}_1} > \boxed{\mathbf{l}_{12}}\right],\boxed{\mathbf{l}_1} > \boxed{\mathbf{l}_{11}},\ \ldots\}\\ \text{CONDS}\left\{\ldots,\begin{bmatrix}\text{LABEL} & \boxed{\mathbf{l}_{11}}\\ \text{DREF} & \boxed{\text{x}}\end{bmatrix},\ \ldots\right\}\end{bmatrix},\ \boxed{2}\left[\boxed{\mathbf{l}_1} > \boxed{\mathbf{l}_{12}}\right],\boxed{\text{x}}\right) := \boxed{\text{x}}$$

$$dref_res\left(\begin{bmatrix} \text{LS}\begin{bmatrix}\text{L-MAX} & \boxed{\mathbf{l}_1}\\ \text{L-MIN} & \boxed{\mathbf{l}_{12}}\end{bmatrix}\\ \text{SUBORD}\{\ldots,\ \boxed{\mathbf{l}_1} \geq \boxed{\mathbf{l}_{12}},\ \ldots\}\end{bmatrix},\ Cond,\boxed{\text{Dref}}\right) := \boxed{\text{Dref}}$$

Note that the first and second clause of *dref_res* do only apply to singular NPs and generalized quantifiers, which contain identity or strict-subordination constraints by the lexical definition of the respective functional categories (see (30), (33)). By contrast, for plural NPs, which are represented as *potentially scope bearing* by a weak subordination constraint as shown in (34), these clauses will fail: the required subordination conditions will not be contained in the SUBORD value of the verb argument.[25] Underspecified as well as disam-biguated plural NPs, characterized by a *weak* subordination constraint in the local UDRS, are captured by the third clause of *dref_res*.

[25] This will be so even if – by the function *pl_dis* to be introduced below – further, disambiguating cons-

Contrary to the first clauses, the variable *Cond*, which is subject to the delay statement on *dref_res*, is not coindexed with a subordination statement in the local SUBORD value. Thus, if no disambiguating constraints are available to determine one of the various possible readings for plural NPs, this argument will remain uninstantiated and the evaluation of the function is blocked. This is what we aimed at for the special concerns of plural underspecification.

If, however, the lexical meaning of the verb determines a particular reading of a plural NP, as e.g. *gather* (see example (10)), the appropriate definition of *dref_res* ensures the correct plural interpretation and relieves the function from its delayed status. This is illustrated in (43). The subject argument is constrained to take a plural DP which is required to be interpreted collectively by stating an appropriate constraint in the SUBORD value of the verb, which characterizes the argument as *not scope bearing.* The function *dref_res* is defined to return the plural discourse referent **X**, defined in the argument UDRS, by coindexation with the third argument place, and moreover, its second argument is instantiated by the identity statement $\mathbf{l}_1 = \mathbf{l}_{12}$. Again the function thereby gets undelayed and by application of the third clause of *dref_res* the discourse referent filling the first argument slot of the verb gets appropriately defined by **X**.

(43)
$$
\left[\text{L}\left[\begin{array}{ll}
\text{CAT}\mid\text{H}\mid\text{SC} & < \text{DP}\left[\begin{array}{ll}\text{CASE } \textit{nom} & \\ \text{NUM } \textit{pl} & \\ \text{UDRS } \boxed{1} & \left[\begin{array}{l} \text{LS}\left[\begin{array}{l}\text{L-MAX } \boxed{l_1}\\ \text{L-MIN } \boxed{l_{12}}\end{array}\right]\\ \text{SUBORD } \{..,\boxed{l_1} \geq \boxed{l_{12}},..\}\\ \text{CONDS } \{..,\left[\begin{array}{l}\text{LABEL } \boxed{l_1}\\ \text{DREF } \boxed{X}\end{array}\right],..\}\end{array}\right]\end{array}\right], \text{PP}\left[\begin{array}{l}\text{PCASE } \textit{loc}\\ \text{UDRS } \boxed{2}\end{array}\right] > \\
\text{UDRS} & \left[\begin{array}{ll} \text{LS}\,[\text{L-MIN } \boxed{1}] & \\ \text{SUBORD} & \{\boxed{l_1} = \boxed{l_{12}}\}\\ \text{CONDS} & \left\{\left[\begin{array}{ll}\text{LABEL} & \boxed{1}\\ \text{REL} & \textit{gather}\\ \text{ARG1} & \textit{dref_res}(\boxed{1},\boxed{l_1}=\boxed{l_{12}},\boxed{X})\\ \text{ARG2} & \textit{dref_res}(\boxed{2},\textit{Cond2},\textit{Dref2})\end{array}\right]\right\}\end{array}\right]
\end{array}\right]\right]
$$

In most cases, however, disambiguating information for the interpretation of plurals comes from various sources of semantic or pragmatic knowledge. Usually it is only provided by previous or – more frequently – subsequent discourse. Thus, we have to define a mechanism for plural disambiguation which may apply at *any* stage of the derivation, to add disambiguating DRS conditions and subordination constraints to the underspecified representation whenever enough information is available to determine a particular plural interpretation. Furthermore, the mechanism for disambiguation will have to trigger the evaluation of the traints for, e.g., a collective or distributive reading are introduced at a later stage of the derivation: the first argument of *dref_res* is coindexed with the UDRS value of an argument in the lexicon entry of the verb. The value of this local UDRS attribute, and with it the SUBORD attribute, will remain unaffected by the introduction of additionally constraining subordination restrictions by the clauses of the Semantics Principle.

delayed function *dref_res*, which then returns the appropriate discourse referent to fill the argument position in the partial DRS of the verb.

We will therefore extend the Semantics Principle to include a function *pl_dis* (*plural disambiguation*), which applies to the phrase's value of UDRS, to render a new value of the same type, *udrs*, which specifies a collective, distributive, or generic reading for a plural discourse referent contained in the original, underspecified representation.
Besides the introduction of disambiguating conditions the individual clauses of *pl_dis* must state constraints which trigger the respective readings, and which are to be satisfied by the preceding context, represented in UDRS. Ideally, these constraints have access to inference modules, including semantic and pragmatic knowledge. We will first state the function *pl_dis* for the different readings and then incorporate the function into the Semantics Principle.

For the collective reading, clause (44) of *pl_dis* must be defined to strengthen the weak subordination relation between the minimal and maximal label of the plural NP to the identity relation. If the constraints which determine a collective interpretation of the plural discourse referent X with identifying label $\mathbf{l}_1$ are satisfied, the subordination restriction $\mathbf{l}_1 = \mathbf{l}_{12}$ is unioned to the original SUBORD value. Note that the function *pl_dis* is fully *monotonic* in that its result is a UDRS which is obtained by only *adding* information to the input values SUBORD and CONDS by union.
As mentioned above, whenever disambiguation of a plural NP takes place, the function *dref_res* must be relieved from its delayed status in order to instantiate the appropriate value in the corresponding argument slot of the verb. We will access the delayed goal *dref_res* with the plural NP's maximal and minimal labels $\mathbf{l}_1$ and $\mathbf{l}_{12}$, define its value by coindexation with $\mathbf{l}_1$'s DREF value $\mathbf{X}$, and instantiate its delayed argument place by the identity constraint $\mathbf{l}_1 = \mathbf{l}_{12}$.
The resulting UDRS for (11) is given in graph notation below.

$$
pl_dis\left(\begin{bmatrix}\text{LS } \boxed{3} \\ \text{SUBORD } \boxed{2}\ \{\ \ldots, \boxed{\text{l}_1} \geq \boxed{\text{l}_{12}}, \ldots\} \\ \text{CONDS } \boxed{1}\ \left\{\ldots, \begin{bmatrix}\text{LABEL } \boxed{\text{l}_1} \\ \text{DREF } \boxed{\text{X}}\end{bmatrix}, \ldots\right\}\end{bmatrix}\right) :=
$$

(44) *constraints triggering a collective reading (of X)* &

$$
\exists \text{ delayed-goal: } dref_res\left(\left[\text{LS}\begin{bmatrix}\text{L-MAX } \boxed{\text{l}_1} \\ \text{L-MIN } \boxed{\text{l}_{12}}\end{bmatrix}\right], \boxed{\text{l}_1} = \boxed{\text{l}_{12}}, \boxed{\text{X}}\right) \ \&
$$

$$
\begin{bmatrix}\text{LS } \boxed{3} \\ \text{SUBORD } \boxed{2} \cup \{\ \boxed{\text{l}_1} = \boxed{\text{l}_{12}}\} \\ \text{CONDS } \boxed{1}\end{bmatrix}
$$

(45)

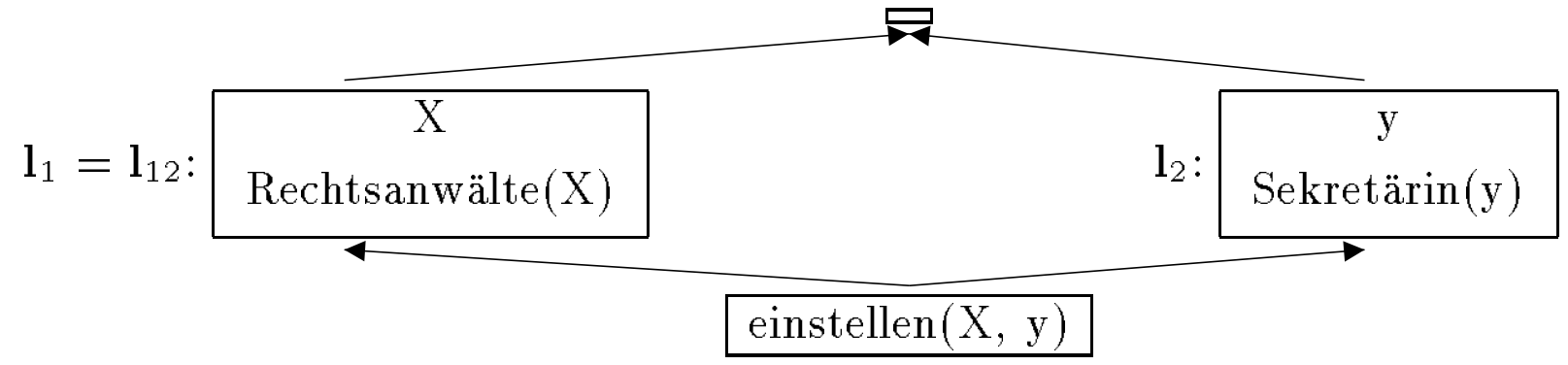

Disambiguation to a distributive reading is obtained in (46) – if appropriate constraints are fulfilled by the preceding context and semantic knowledge – by adding quantificational conditions for the distributive reading to the original value of CONDS. The restrictor $\mathbf{l}_{11}$ states the distribution condition $\mathbf{x} \in \mathbf{X}$ and the nuclear scope is identified by the minimal label $\mathbf{l}_{12}$. Moreover, the (strong) subordination of restrictor and scope is defined in SUBORD. Again, the delayed goal *dref_res* for the definition of discourse referent filling the argument slot of the verb is defined – now by the individual type referent $\mathbf{x}$ – and is un-delayed by instantiation of its second argument position. The resulting UDRS is displayed in (47).

$$pl_dis\left(\begin{bmatrix} \text{LS } \boxed{3} \\ \text{SUBORD } \boxed{2}\ \{\ldots, \boxed{l_1} \geq \boxed{l_{12}}, \ldots\} \\ \text{CONDS } \boxed{1}\ \left\{\ldots, \begin{bmatrix} \text{LABEL } \boxed{l_1} \\ \text{DREF } \boxed{X} \end{bmatrix}, \ldots\right\} \end{bmatrix}\right) :=$$

(46)

$$\begin{array}{l} \textit{constraints triggering a distributive reading (of X)}\ \& \\ \exists \text{ delayed-goal: } dref_res\left(\begin{bmatrix} \text{LS} \begin{bmatrix} \text{L-MAX } \boxed{l_1} \\ \text{L-MIN } \boxed{l_{12}} \end{bmatrix} \end{bmatrix}, \boxed{l_1} > \boxed{l_{12}}, \boxed{x}\right)\ \& \\ \begin{bmatrix} \text{LS } \boxed{3} \\ \text{SUBORD } \boxed{2} \cup \{\ \boxed{l_1} > \boxed{l_{11}}, \boxed{l_1} > \boxed{l_{12}}\} \\ \text{CONDS } \boxed{1} \cup \left\{ \begin{bmatrix} \text{LABEL } \boxed{l_1} \\ \text{REL } \Rightarrow \\ \text{RES } \boxed{l_{11}} \\ \text{SCOPE } \boxed{l_{12}} \end{bmatrix}, \begin{bmatrix} \text{LABEL } \boxed{l_{11}} \\ \text{DREF } \boxed{x} \\ \text{REL } \in \\ \text{ARG1 } \boxed{x} \\ \text{ARG2 } \boxed{X} \end{bmatrix} \right\} \end{bmatrix} \end{array}$$

(47)

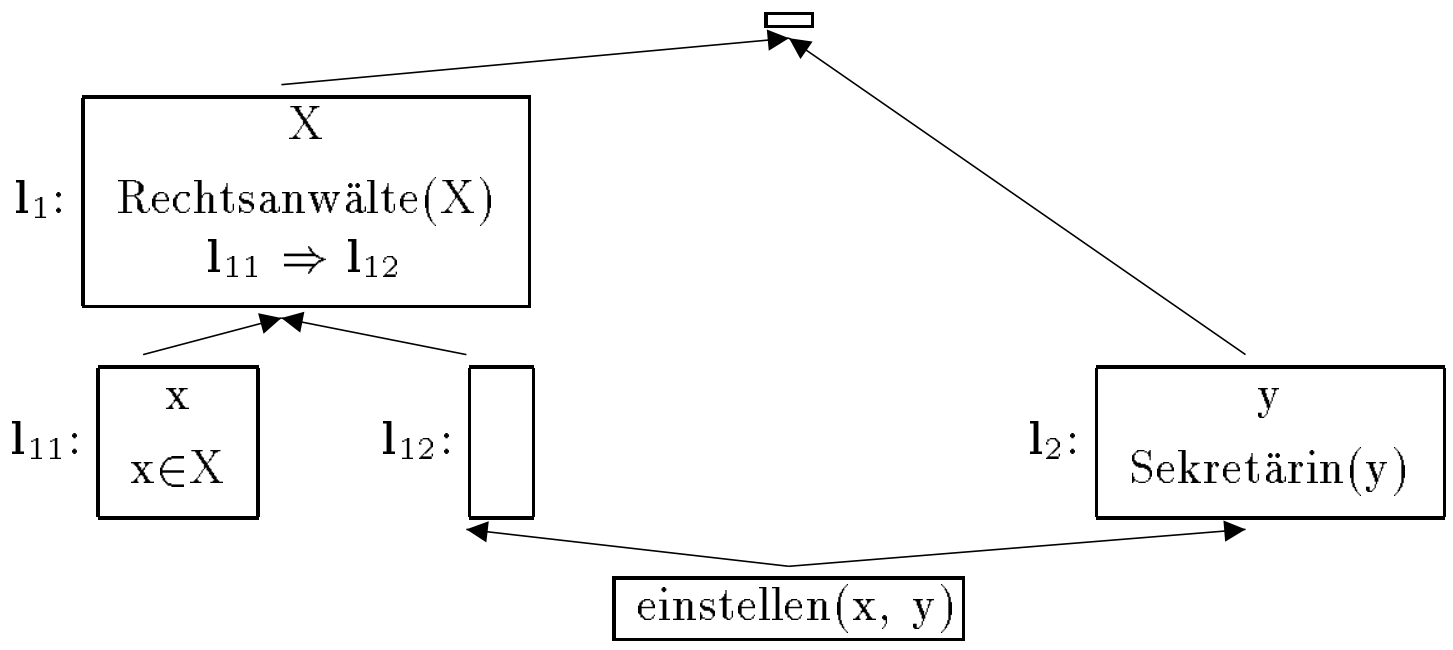

The definition of *pl_dis* for generic and cumulative readings is straightforward and will not be made explicit here. Finally, a trivial clause for *pl_dis* is defined as the identity function in case no disambiguating information is available.

We now complete the Semantics Principle by the Principle for Plural Disambiguation (VII): At any stage of the analysis the function *pl_dis* may apply to the phrase's UDRS value, which is defined by the principles for UDRS construction and the scoping principles stated above.

Depending on the preceding context, represented in the UDRS value, and supplemented by general semantic and/or pragmatic knowledge, *pl_dis monotonically* redefines the phrase's UDRS value if disambiguating constraints for a specific plural reading can be determined.

It is only one step further then to state a principle governing anaphora resolution in a similar way, i.e. by accessing inference modules using semantic and pragmatic knowledge, in order to impose constraints on the interpretation of anaphors in underspecified discourse representations. But this is a big step, to be reserved for future research.

**Semantics Principle:**[26]
**(I) & (II) & ... & (VII) Plural Disambiguation**

(48)

$$\left[\begin{array}{l} \text{SYM} \mid \text{LOC} \mid \text{UDRS} \;\; \textit{pl_dis} \left( \left[\begin{array}{ll} \text{LS} & \boxed{5} \\ \text{SUBORD} & \ldots \cup \boxed{3} \cup \boxed{4} \\ \text{CONDS} & \boxed{1} \cup \boxed{2} \end{array}\right] \right) \\ \textit{head-comp-struc} \end{array}\right]$$

C-DTR: $$\left[ \text{SYM} \mid \text{LOC} \mid \text{UDRS} \left[\begin{array}{ll} \text{SUBORD} & \boxed{4} \\ \text{CONDS} & \boxed{2} \end{array}\right] \right]$$

H-DTR: $$\left[ \text{SYM} \mid \text{LOC} \mid \text{UDRS} \left[\begin{array}{ll} \text{LS} & \boxed{5} \\ \text{SUBORD} & \boxed{3} \\ \text{CONDS} & \boxed{1} \end{array}\right] \right]$$

Many questions arise once we include the function *pl_dis* into the Semantics Principle. First of all, its status differs essentially from the remaining clauses of the Semantics Principle: While the clauses (I) – (IV) define the core mechanism for UDRS construction, and clauses (V) and (VI) govern the syntactically determined scoping conditions for quantified arguments, the clause (VII) for plural disambiguation provides a powerful interface to contextual and pragmatic reasoning modules.
It has to be carefully considered how such a powerful device can be appropriately restricted, for it is evident that for reasons of efficency inferencing modules for disambiguation should only be accessed if there is a sufficient amount of 'new' and 'relevant' information available which (i) provides new criteria for disambiguation or (ii) triggers new sources for ambiguities. It may therefore be advisable to restrict the application of *pl_dis* to the sentence level. Plural disambiguation can then only take place when the representation of a complete new sentence is available for contextual reasoning. It might also be useful to consider insights from the theory of incremental interpretation in order to develop a promising controlling strategy for this device.

A further issue, which is discussed in the field of incremental interpretation and which is also interesting for the present account, is the issue of generating disambiguating *hypotheses*.[27] It may be argued that in incremental interpretation there is seldom enough clear-cut evidence for one or the other reading of a plural NP, while on the other hand disambiguation may be led by strong hypotheses favouring a particular reading. We may therefore decide to trigger plural disambiguation if there is *sufficiently* strong evidence for a particular reading, and allow for revision of the semantic representation in case the hypothesis gets falsified by subsequent discourse (see e.g. so-called *jungle paths*, the semantic equivalent to *garden paths* [Barwise]).
Since the UDRS construction in general, and especially the function *pl_dis* are fully monotonic, it should in principle be possible to deal with revision of hypothetical assumptions if we get hold of the triggering hypotheses as 'choice points'. Again, we cannot even touch these interesting issues.

[26] We again indicate by dots the subordination restrictions which are defined by the clauses (IV) – (VI) of the Semantics Principle.

[27] David Milward, p.c., at a workshop on incrementality and underspecification at the European Summer School, Copenhagen, 1994.

At long last we have to reconsider the scoping principles for the case of underspecified plural NPs. The Quantifier Scope Principle (V) and Complement Scope Principle (VI) were defined to apply to *both* generalized quantifiers and *potentially scope bearing* plural NPs. This was motivated by the assumption that plural NPs – if they get a quantificational (distributive or generic) reading – are subject to the Scope Principle of [Frey]. Yet, if instead a non-quantificational, i.e. collective reading is called for, the plural NP may get arbitrarily wide scope. Thus the introduction of scoping constraints for plural NPs must be restricted to plural NPs which will *in fact* be disambiguated as *scope bearing* elements. The main problem here is that plural disambiguation may take place rather late in subsequent discourse, while the syntactic constraints for quantificational scope can only be determined locally.

We have seen above, when we defined the clauses for plural disambiguation, that there is a way to distinguish *actually scope bearing* from *finally not scope bearing* plurals in terms of their minimal and maximal labels: they are resolved to distinct or identical values, respectively. Instead of introducing scoping conditions for *potentially* scope bearing plurals, then, the Quantifier Scope Principle (V) and Complement Scope Principle (VI) introduce *conditionalized* subordination restrictions instead of the ones stated below:

If the maximal and minimal labels $\mathbf{l}_{max_\beta}$ and $\mathbf{l}_{min_\beta}$ of a (potentially) scope bearing element $\beta$ are distinct,

**(i)** the maximal label of $\beta$ is subordinated to the label $\mathbf{l}_{max}$ identifying its local domain:
$\mathbf{l}_{max_\beta} > \mathbf{l}_{min_\beta} \Rightarrow \mathbf{l}_{loc_domain} \geq \mathbf{l}_{max}$

**(ii)** every (potentially) scope bearing element $\alpha$ that is c-commanded by $\beta$, if the minimal and maximal labels of $\alpha$ bear distinct values, the maximal label of $\alpha$ is subordinated to the minimal label of $\beta$:
$\mathbf{l}_{max_\beta} > \mathbf{l}_{min_\beta} \Rightarrow (\ \mathbf{l}_{max_\alpha} > \mathbf{l}_{min_\alpha} \Rightarrow \mathbf{l}_{min_\beta} \geq \mathbf{l}_{max_\alpha}\ )$

We can now state the (revised) Semantics Principle in full shape:

**Semantics Principle: (I) & (II) & (III) & (IV) &**
**(V) Quantifier Scoope Principle & (VI) Complement Scope Principle & (VII)**

$$\left[ .. \begin{bmatrix} \text{LOC} \mid \text{UDRS} \;\; pl_dis \left( \begin{bmatrix} \text{LS } \boxed{5} \\ \text{SUBORD } \boxed{10} \cup \boxed{11} \cup \{ \boxed{l_{min}} \geq \boxed{l_{verb}} \} \cup \boxed{3} \cup \boxed{4} \\ \text{CONDS } \boxed{1} \cup \boxed{2} \end{bmatrix} \right) \\ \text{NLOC} \mid \text{INH} \mid \text{SLASH } \boxed{9} \end{bmatrix} \right]_{head-comp-struc}$$

(49) C-DTR H-DTR

$$\left[ ..\boxed{7}\,\text{LOC} \begin{bmatrix} \text{CAT} \mid \text{HEAD} \;\; quant \vee plural \\ \text{UDRS} \begin{bmatrix} \text{LS} \begin{bmatrix} \text{L-MAX } \boxed{l_{quant}} \\ \text{L-MIN } \boxed{l_{min}} \end{bmatrix} \\ \text{SUBORD } \boxed{4} \\ \text{CONDS } \boxed{2} \end{bmatrix} \end{bmatrix} \right] \qquad \left[ .. \begin{bmatrix} \text{CAT} \mid \text{HEAD} \mid \text{SUBCAT } \boxed{6} \oplus \boxed{7} \oplus \boxed{8} \\ \text{UDRS} \begin{bmatrix} \text{LS } \boxed{5} \begin{bmatrix} \text{L-MAX } \boxed{l_{max}} \\ \text{L-MIN } \boxed{l_{verb}} \end{bmatrix} \\ \text{SUBORD } \boxed{3} \\ \text{CONDS } \boxed{1} \end{bmatrix} \end{bmatrix} \right]$$

- $\boxed{10}$ is the set of conditions of the form:
  $\boxed{l_{quant}} > \boxed{l_{min}} \Rightarrow (\boxed{l_{\alpha_{max}}} > \boxed{l_{\alpha_{min}}} \Rightarrow \boxed{l_{min}} \geq \boxed{l_{\alpha_{max}}})$
  where $\boxed{l_{\alpha_{max}}}$ and $\boxed{l_{\alpha_{min}}}$ are the maximal and minimal labels of every element $\alpha$ in $\boxed{8}$ s.th. $\alpha$ is of type *quant* $\vee$ *plural* and the LOC value of $\alpha$ is not contained in SLASH $\boxed{9}$

- $\boxed{11} = \{ \boxed{l_{quant}} > \boxed{l_{min}} \Rightarrow \boxed{l_{max}} \geq \boxed{l_{quant}} \}$

# 7 Conclusion and further Perspectives

A constraint based semantic formalism for HPSG has been presented to replace the standard approach of HPSG to semantics. It has been pointed out that the new formalism comes closer to a principle based construction of semantic structure and, therefore, is more in the spirit of HPSG philosophy than its standard approach. Furthermore the new formalism overcomes a number of shortcomings of the standard approach in a natural way.

In particular, we presented an HPSG grammar for German that defines a syntax-semantics interface for the construction of U(nderspecified) D(iscourse) R(epresentation) S(tructure)s. The construction is guided by general principles, which clearly identify the interaction between the modules, i.e. the "interface" between syntax and semantics. In the fragment we defined underspecificied representations for quantificational structures and plural NPs. The principles governing the interaction of syntax and semantics specify scoping relations for quantifiers and quantificational readings of plural NPs, where syntactic constraints of word order restrict the set of possible readings.

In addition to the syntax/semantics interface the Semantics Principle developped in this paper also defines a clear interface to contextual and pragmatic knowledge. This interface allows reasoning modules to interact with semantics construction. The approach taken here can, therefore, be generalized to disambiguation problems other than the collective/distributive ambiguity as well as to anaphora resolution. A further issue to which the present account is directly related is incremental interpretation.

## Literatur

[Abb/Maienborn] Abb, B. / Maienborn, C. (1994): "Adjuncts in HPSG", in: Trost, H. (ed): *KONVENS '94. Verarbeitung natürlicher Sprache.* Informatik Xpress 6, Springer-Verlag, Berlin, 13–22.

[Alshawi] Alshawi, Hiyan (1990): "Resolving Quasi Logical Forms", in *Computational Linguistics*, Vol. 16, No. 3.

[Alshawi/Crouch] Alshawi, H. / Crouch, R. (1992) "Monotonic Semantic Interpretation". in: *Proceedings of the 30th ACL*, University of Delaware, 32–39.

[Barwise] Barwise, J. (197): Noun Phrases, Generalized Quantifiers and Anaphors. in: Gardenfors, P. (ed): *Generalized Quantifiers*, Dordrecht, Reidel.

[Cooper] Cooper, R. (1983) *Quantification and Syntactic Theory.* Reidel, Dordrecht, 1–29.

[Doerre/Dorna] Dörre, J., Dorna, M. (1993) "CUF – A Formalism for Linguistic Knowledge Representation." in: Dörre, J. (ed): *Computational Aspects of Constraint-Based Linguistic Description I.* ESPRIT Basic Research Action BR-6852 (DYANA-2), Deliverable R1.2.A.

[Fenstad et. al.] Fenstad, J.E. / Halvorsen, P.-K. / van Benthem, J. (1987) *Situations, Language and Logic*, Reidel Publishing Company.

[Frank] Frank, A. (1994): "Verb Second by Underspecification", in: Trost, H. (ed): *KONVENS '94. Verarbeitung natürlicher Sprache.* Informatik Xpress 6, Springer-Verlag, Berlin, 121–130.

[Frank/Reyle] Frank, A. / Reyle, U. (1992): "How to Cope with Scrambling and Scope", in: Görz, G. (ed.) *KONVENS '92.* Reihe Informatik aktuell, Springer-Verlag, Berlin, 178–187.

[Frey] Frey, Werner (1993): *Syntaktische Bedingungen für die semantische Interpretation*, Studia Grammatica Bd. XXXV, Akademie Verlag, Berlin.

[Frey/Tappe] Frey, Werner / Tappe Thilo (1992): *Grundlagen eines GB-Fragments für das Deutsche*, appears as *Arbeitspapier des Sonderforschungsbereichs 340*, Stuttgart.

[Grimshaw] Grimshaw, J. (1991): "Extended Projections", ms, Brandeis University, Mass.

[Haider] Haider, H. (1993): *Deutsche Syntax - Generativ. Vorstudien zur Theorie einer projektiven Grammatik*, Narr, Tübingen.

[Hoehle] Hoehle, Tilman (1982): "Explikation fuer 'normale' Betonung und 'normale Wortstellung'. " in: W. Abraham (ed.): *Satzglieder im Deutschen. Vorschlaege zur syntaktischen, semantischen und pragmatischen Fundierung.* Tuebingen, Narr, 75–153.

[Kamp/Reyle] Kamp, H. / Reyle, U. (1993): *From Discourse to Logic*, Reidel, Dordrecht.

[Nerbonne] Nerbonne, J. (1992), "A Feature Base Syntax/Semantics Interface", ms. Saarbruecken.

[Netter] Netter, K. (1994): "Towards a Theory of Functional Heads: German Nominal Phrases". to appear in: Nerbonne, J. / Netter, K. / Pollard, C. (eds.) *German in Head Driven Phrase Structure Grammar.* CSLI Lecture Notes, Chicago UP.

[Peters/vanDeemter95] Peters, S. / van Deemter, C.J. (eds.) (1995): "Semantic Ambiguity and Underspecification" (tentative title), to appear in: CSLI Lecture Notes.

[Pollard/Sag87] Pollard, C. / Sag, I.A. (1987): *Information-Based Syntax and Semantics*, CSLI Lecture Notes Series 13.

[Pollard/Sag] Pollard, C. / Sag, I.A. (1994): *Head-Driven Phrase Structure Grammar*, Chicago: University of Chicago Press and Stanford: CSLI Publications.

[Reyle 92] Reyle, Uwe (1993): "Dealing with Ambiguities by Underspecification: A First Order Calculus for Unscoped Representations", in: *Proceedings of the Eighth Amsterdam Colloquium*, Amsterdam.

[Reyle 93] Reyle, Uwe (1993): "Dealing with Ambiguities by Underspecification: Construction, Representation and Deduction", in: *Jounal of Semantics*, 10(2).

[Reyle 94] Reyle, Uwe (1994): "Monotonic Disambiguation and Plural Pronoun Resolution", ms. Universität Stuttgart, submitted to: Peters, S. / van Deemter, C.J. (eds.) (1995): "Semantic Ambiguity and Underspecification" (tentative title), to appear in: CSLI Lecture Notes.

[Schubert/Pelletier] Schubert, L.K. / Pelletier, F.J. (1982), "From English to Logic: Context-Free Computation of Conventional Logic Translations", in: *Journal of the Association for Computational Linguistics.*